\documentclass[]{bytedance_seed}

\usepackage[toc,page,header]{appendix}

\usepackage{minitoc}

\usepackage{dblfloatfix}
\usepackage{minted}
\usepackage{xspace}
\usepackage{amsmath,amssymb}
\usepackage{makecell}

\newmintinline{python}{python3}
\newminted{python}{frame=single,framesep=1mm,fontsize=\scriptsize,python3,linenos}
\newmintinline{md}{markdown}

\newminted{mdcode}{
    style=colorful,
    frame=single,
    framesep=1mm,
    fontsize=\scriptsize,
    linenos,
    breaklines,
    tabsize=2
}

\newcommand{\ours}{CUDA Agent\xspace}

\usepackage{tikz}
\usetikzlibrary{calc}
\usepackage{xcolor}

\definecolor{gensicolor}{RGB}{220,245,84}  
\definecolor{seedcolor}{RGB}{120,230,220}  

\newcommand{\gensifont}[1]{%
  \tikz[baseline=(text.base)]{
    \node[inner sep=0pt, outer sep=0pt] (text) {#1};  
    \fill[gensicolor]
      (text.south west) rectangle
      ($(text.south east)!0.45!(text.north east)$);
    \node[inner sep=0pt, outer sep=0pt] {#1};
  }%
}

\title{ \gensifont{CUDA Agent}: Large-Scale Agentic RL \\ for High-Performance CUDA Kernel Generation }

\author[1,2,3*]{Weinan Dai}
\author[1,2,3*]{Hanlin Wu}
\author[1,2,3]{Qiying Yu}
\author[1,2,3]{Huan-ang Gao} 
\author[1]{\\Jiahao Li}
\author[1]{Chengquan Jiang}
\author[1]{Weiqiang Lou}
\author[1]{Yufan Song}
\author[1,2,3]{Hongli Yu}
\author[1,3]{\\Jiaze Chen}
\author[2,3]{Wei-Ying Ma}
\author[2,3]{Ya-Qin Zhang}
\author[2,3]{Jingjing Liu}
\author[1,3]{Mingxuan Wang}
\author[1]{Xin Liu}
\author[2,3\dagger]{Hao Zhou}

\affiliation[1]{ByteDance Seed}
\affiliation[2]{Institute for AI Industry Research (AIR), Tsinghua University\\}

\affiliation[3]{SIA-Lab of Tsinghua AIR and ByteDance Seed}

\contribution[*]{Equal contributions}
\contribution[\dagger]{Corresponding Authors}

\date{\today}
\correspondence{\email{zhouhao@air.tsinghua.edu.cn}}

\checkdata[Project Page]{\url{https://cuda-agent.github.io/}}

\begin{document}
\abstract{
GPU kernel optimization is fundamental to modern deep learning but remains a highly specialized task requiring deep hardware expertise. Despite strong performance in general programming, large language models (LLMs) remain uncompetitive with compiler-based systems such as \texttt{torch.compile} for CUDA kernel generation. Existing CUDA code generation approaches either rely on training-free refinement or fine-tune models within fixed multi-turn execution-feedback loops, while both paradigms fail to fundamentally improve the model’s intrinsic CUDA optimization ability, resulting in limited performance gains. We present \ours, a large-scale agentic reinforcement learning system that develops CUDA kernel expertise through three components: a scalable data synthesis pipeline, a skill-augmented CUDA development environment with automated verification and profiling to provide reliable reward signals, and RL algorithmic techniques enabling stable training. \ours achieves state-of-the-art results on KernelBench, delivering 100\%, 100\%, and 92\% faster rate over \texttt{torch.compile} on KernelBench Level-1, Level-2, and Level-3 splits, outperforming the strongest proprietary models \textit{e.g.} Claude Opus 4.5 and Gemini 3 Pro by about 40\% on the hardest Level-3 setting. 
}

\maketitle


\section{Introduction}
\begin{figure*}[t]
  \begin{center}
    \centerline{\includegraphics[width=\linewidth]{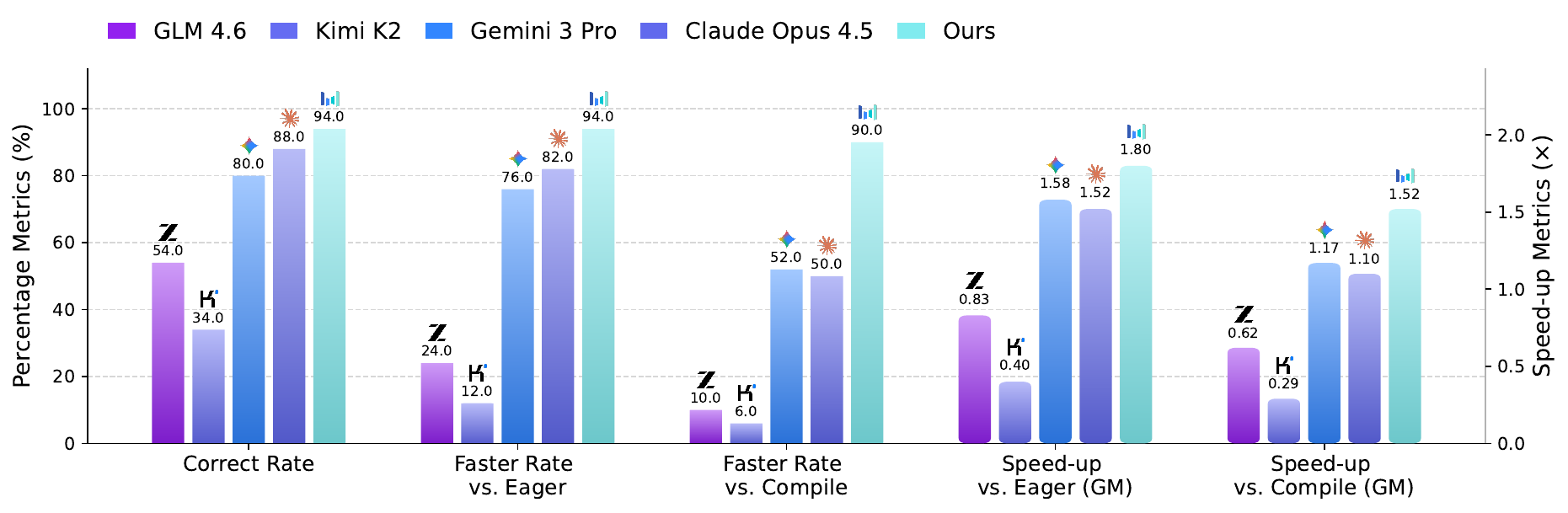}}
    \label{fig:benchmark_chart}
  \end{center}
  \vskip -0.3in
\end{figure*}
GPU kernels are a foundational component of modern deep learning infrastructure \citep{ouyang2025kernelbench, lange2025sakanacuda}, with NVIDIA’s CUDA architecture currently dominating the AI hardware ecosystem. Despite its widespread adoption, the development and optimization of high-performance CUDA kernels remain highly challenging, which requires a deep understanding of GPU microarchitectural features~\cite{NVIDIA_H100}, and sophisticated profiling toolkits \citep{baronio2025kevin}.

Although Large Language Models (LLMs) have demonstrated \emph{human-comparable} proficiency in general software development tasks~\cite{swe-agent}, existing CUDA kernel generation approaches \citep{baronio2025kevin, lange2025sakanacuda} still remain \emph{uncompetitive} even with automatic optimization tools such as \verb|torch.compile| in CUDA kernel code generation~\cite{ouyang2025kernelbench}, let alone human experts.

One line of prior work on CUDA code generation focuses on designing training-free workflows~\cite{dong2025stark,gong2025largesmalltransferringcuda,guo2025evoengineer}, which rely on hand-designed refinement heuristics guided by execution feedback. However, these methods do not remedy the fundamental lack of CUDA-coding abilities in the base models, causing performance gains to be significantly capped by the model’s intrinsic capabilities. Another line of research~\cite{baronio2025kevin,li2025cudaL1,concur, anonymous2026mastering} attempts to fine tune base models within a fixed multi-turn refinement loop driven by code execution feedback. However, such methods waste context length by including all previous solutions and constrain the agent's autonomy to learn debugging, search, and profiling strategies.

To overcome these limitations, we introduce \ours, a large-scale CUDA agent training system that systematically enhances the base model’s CUDA kernel coding capabilities by making contributions in three complementary dimensions: data, agent environment, and reinforcement learning (RL) algorithmic techniques. 

To support large-scale reinforcement learning, we design a scalable data synthesis pipeline that generates training problems spanning a wide range of difficulty levels, enabling effective curriculum-based RL training.
For the agent environment, we adopt the agent skills paradigm~\citep{anthropic2025equipping}, equipping the model with a structured specification that formalizes the standard workflow for writing, validating and optimizing CUDA kernels, together with automated test and profiling scripts for execution-based feedback. Notably, we implement rigorous correctness and performance tests, along with system-level permission isolation, to prevent reward hacking and ensure accurate reward signals.
As for RL algorithmic improvement, we analyze the sources of instability in RL optimization and propose a multi-stage warm-up strategy for both the actor and critic models, enabling stable training of large language models for 150 steps.

Integrating these components, \ours successfully scales to a context length of 128k tokens and supports up to 200 interaction turns, achieving state-of-the-art performance. Specifically, \ours achieves speedups of \textbf{100}\%, \textbf{100}\%, and \textbf{92}\% over \texttt{torch.compile} on the Level-1, Level-2, and Level-3 splits of KernelBench~\citep{ouyang2025kernelbench}, outperforming advanced proprietary models such as Claude Opus 4.5 and Gemini 3 Pro\footnote{We found that models in the ChatGPT-5 series (5, 5.1, and 5.2) were not amenable to evaluation on this task, as they consistently declined to respond to CUDA-related prompts.} by approximately \textbf{40}\% in the Level-3 split.

Our contributions can be summarized as follows:
\begin{itemize}
    \item We introduce \ours, a large-scale agentic reinforcement learning system for automatic CUDA kernel generation, which systematically improves the base model’s CUDA coding and optimization abilities through scalable data synthesis, a skill-augmented CUDA kernel development environment, and RL techniques designed for stable long-context, multi-turn agentic training.
    \item \ours achieves state-of-the-art performance on KernelBench, outperforming \texttt{torch.compile} in the majority of test cases and delivering the largest speedups across multiple difficulty levels. These results establish LLM-based kernel generation as a competitive—and often superior—alternative to traditional compiler-driven kernel optimization.
\end{itemize}
\begin{figure*}[t!]
  \vskip -0.1in
  \begin{center}
    \centerline{\includegraphics[width=0.78\linewidth]{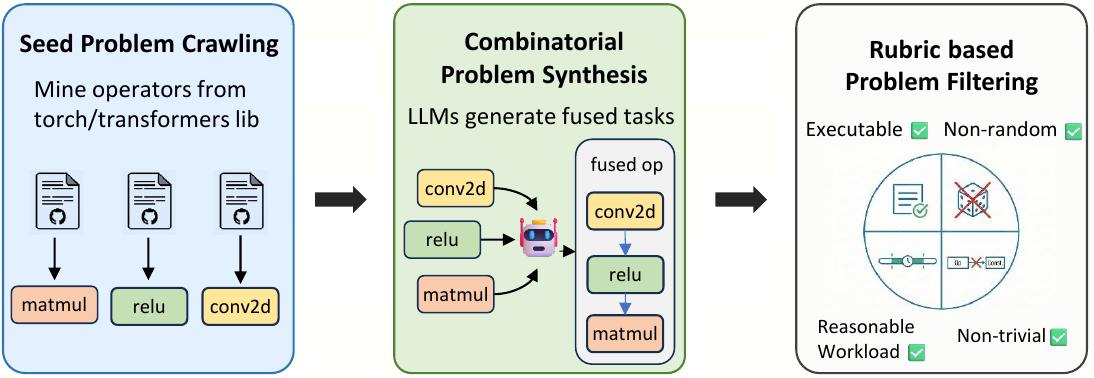}}
    \caption{
    \textbf{Overview of the three-stage data collection pipeline.} We first crawl seed operators from PyTorch and Transformer libraries to build a repository of fundamental computational primitives. Next, an LLM performs combinatorial synthesis to generate fused, multi-operator tasks. Finally, a rubric-based filtering stage retains only executable, deterministic, non-trivial problems with reasonable workloads to ensure data quality and reliable evaluation.
    }
    \label{fig:data-collection-pipeline}
  \end{center}
  \vskip -0.3in
\end{figure*}
\section{Related Works}

\subsection{Training Free Systems for Kernel Generation}

Several recent works explore designing  with explicit search to address the irregular landscape of kernel optimization.
STARK~\cite{dong2025stark} employs a strategic team of agents with planning, coding, and debugging roles that explore a tree-structured search space on KernelBench, iteratively refining kernels using compilation, correctness checks, and timing feedback.
ReGraphT~\cite{gong2025largesmalltransferringcuda} presents a training-free, retrieval-augmented framework that distills CUDA optimization trajectories from large language models into a reasoning graph, which is then searched via Monte Carlo Graph Search to guide smaller models toward competitive performance.
EvoEngineer~\cite{guo2025evoengineer} formulates CUDA kernel optimization as a constrained code evolution problem and applies an LLM-driven evolutionary loop that iteratively edits and validates kernels to improve performance while preserving correctness.
CudaForge~\cite{zhang2025cudaforge} introduces a training-free two-agent system where a Judge uses Nsight Compute profiling and hardware specifications to diagnose bottlenecks and provide targeted optimization feedback to a Coder, achieving consistent speedups across GPUs.

Although these works demonstrate considerable performance gains, they heavily rely on the base model's CUDA coding capability. Furthermore, those test-time scaling approaches are also orthogonal and could be applied to our model.

\subsection{Fine-tuning LLM for Kernel Generation}
In parallel, another line of research seeks to improve CUDA kernel generation by training base models via supervised fine-tuning or reinforcement learning.
Kevin~\cite{baronio2025kevin} introduces a multi-turn reinforcement learning framework for CUDA kernels that models the iterative developer workflow, achieving notable gains in both correctness and performance over its QwQ-32B base model on KernelBench.
CUDA-L1~\cite{li2025cudaL1} proposes a contrastive reinforcement learning framework that evaluates multiple CUDA kernel variants using execution-based rewards. However, their training and evaluation are conducted on the same KernelBench dataset without a proper train–test split. This data leakage renders the reported results not directly comparable to ours.
ConCuR~\cite{concur} synthesize and curate CUDA kernels with reasoning traces, and uses the resulting dataset to fine-tune QwQ-32B into KernelCoder, achieving performance improvment among open-source models.

Overall, these approaches remain fundamentally constrained by the scarcity of high-quality training data, limited training scale, and hand-designed optimization loops, which collectively cap their performance improvements. A more detailed analysis and comparison of these related works can be found in Appendix~\ref{appd:concurrent_works}.

\begin{figure*}[t!]
  \begin{center}
    \centerline{\includegraphics[width=0.8\linewidth]{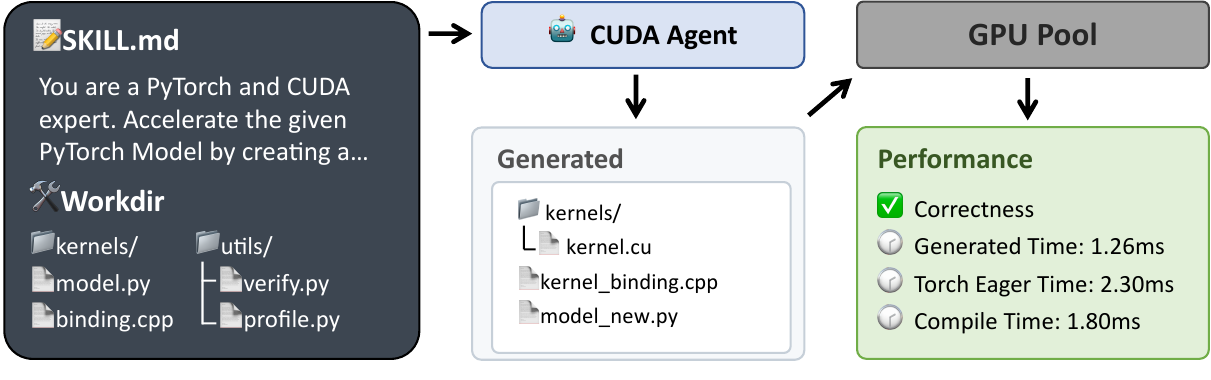}}
    \caption{
    \textbf{Overview of the agent loop.} 
    }
     \label{fig:agent_loop}
  \end{center}
  \vskip -0.3in
\end{figure*}

\section{Method}

Our large-scale agentic reinforcement learning system \ours is built upon three key components: 1) a scalable data collection pipeline; 2) a skill-integrated and non-hackable training environment with robust reward scheduling designing; and 3) proposed reinforcement learning algorithmic techniques for stable training.
Each component plays a crucial role in achieving the final strong empirical results.

\subsection{Scalable Training Data Synthesis Pipeline}
\label{sec:data_pipeline}

The scarcity of high-performance CUDA kernels creates a significant bottleneck for supervised fine-tuning, as manual implementation of expert-level reference code is prohibitively expensive. To overcome this, we employ reinforcement learning (RL) for training \ours. RL necessitates a vast and diverse corpus of reference operators implemented in PyTorch to serve as training tasks. Since existing public datasets lack the requisite diversity and scale, we develop a \textbf{scalable data collection pipeline} that systematically expands the task space through seed problem crawling, LLM-based combinatorial synthesis, and rigorous execution-based filtering (Figure~\ref{fig:data-collection-pipeline}).

The key observation for our combination based data synthesis is composing multiple operators into fused tasks yields valuable new CUDA-agent training problem. This is because the combined problem is often not equivalent to trivially optimizing each operator in isolation and then chaining them. Fusion reshapes the optimization landscape by avoiding intermediate global-memory materialization, coupling stages through shared register/SMEM/occupancy constraints, and requiring a unified parallel mapping and data layout that may favor downstream consumption.

\textbf{Seed Problem Crawling} First, we mine reference operator implemented in PyTorch from the \texttt{torch} and \texttt{transformers} libraries, establishing a comprehensive seed problem set. Each operator is represented as a Python class with initialization and forward methods. These operator classes are widely used and well-maintained. We therefore exclude individually maintained repositories that lack sufficient code quality.

\textbf{Combinatorial Problem Construction} Next, to expand the dataset and introduce higher complexity, we utilize LLMs to synthesize aggregated operators. Specifically, the LLM is prompted to sample no more than 5 operator classes from the \texttt{torch} library. The sampled operator classes are composed sequentially by stacking them into a single computational layer. We do not sample operator classes from the \texttt{transformers} library, as these operators are typically higher-level modules that already encapsulate multiple primitive operations.

\textbf{Problems Filtering} Finally, we implement a rigorous data selection process to filter out problems that are either too easy or too difficult, based on execution feedback. We validate each operator against four criteria: (1) The operator must execute successfully in both Eager and Compile modes. 
(2) To ensure reproducibility, we exclude operators with inherent stochasticity. 
(3) For anti-hacking, we verify that outputs for different inputs are neither constant values nor numerically indistinguishable. 
(4) To filter out trivial or excessively heavy tasks, we restrict the execution time in eager mode to the range of 1\,ms to 100\,ms. Furthermore, we exclude operators that exhibit high similarity to KernelBench test cases; the similarity distribution is provided in \cref{appd:data_details}.

In total, the filtered synthesized training dataset contains 6,000 samples, forming CUDA-Agent-Ops-6K\footnote{\url{https://huggingface.co/datasets/BytedTsinghua-SIA/CUDA-Agent-Ops-6K}}, a curated operator-level dataset for training CUDA-capable agents. Additional details on data format, contamination analysis, and dataset composition are provided in \cref{appd:data_details}.

\subsection{Skill-Integrated Agent Loop}

\label{sec:agent_loop}

\textbf{Agent Loop} As CUDA kernel coding naturally arises as a subtask of coding agents, we design our agent loop to align with the widely adopted OpenHands framework~\citep{wang2024openhands} to ensure the generalizability. The LLM is provided with a standard suite of shell utilities—\texttt{BashTool}, \texttt{GlobTool}, \texttt{MultiEditTool}, and \texttt{TodoWriteTool}—to fully support CUDA coding development (see \cref{appd:agent_loop_details} for the full list). On top of this, our agent loop (visualized in \cref{fig:agent_loop}) follows the ReAct-style paradigm~\citep{yao2022react}, interleaving reasoning, action execution, and observation to enable iterative coding, debugging, and performance optimization.

\textbf{CUDA Coding Skill} Inspired by the idea of Agent Skills \citep{anthropic2025equipping}, we deliberately put the CUDA coding specific instructions and tools (\emph{e.g.} the profiling tool to compare the performance of generated kernel against \texttt{torch.compile}) as Agent Skills format. We also design a CUDA kernel coding specific instructions \texttt{SKILL.md} formulating the standard process to optimize the CUDA kernel (original text can be found in \cref{appd:agent_loop_details}):
\begin{enumerate}
    \item Analyze the performance of the native PyTorch implementation using the provided \texttt{profile.py} script. This step identifies performance bottlenecks and optimization opportunities, such as excessive kernel launches and suboptimal memory access patterns.
    \item Implement custom CUDA operators by rewriting the model in \verb|model_new.py| and developing the corresponding CUDA kernel source files and binding code, targeting the performance-critical operators identified in the analysis stage.
    \item Compile and evaluate the optimized model in the provided GPU sandbox environment, and iteratively refine the CUDA kernel implementations until both numerical correctness and performance requirements are met.
    \item Repeat the optimization process starting from Step 2 until the final implementation achieves at least a 5\% speedup over the \texttt{torch.compile} baseline while passing all numerical correctness checks.
\end{enumerate}
\begin{figure*}[htbp]
  \begin{center}
    \centerline{\includegraphics[width=0.9\linewidth]{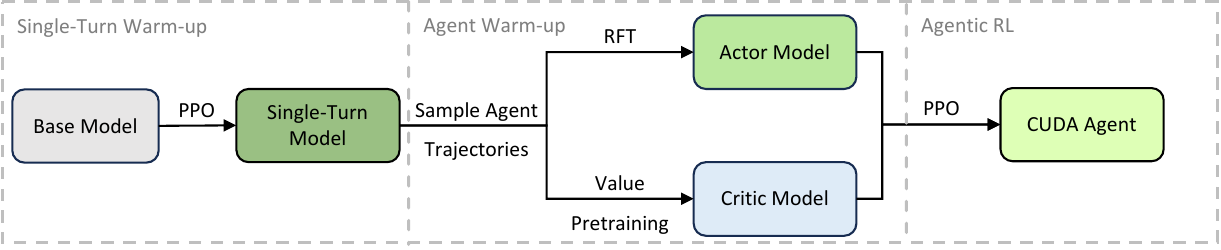}}
    \caption{
    \textbf{Overview of training pipeline.} Following a single-turn RL warm-up stage, the sampled trajectories are used to initialize actor model and critic model before agentic RL stage.
    }
    \label{fig:training-pipeline}
  \end{center}
  \vskip -0.3in
\end{figure*}

\textbf{Robust Reward Scheduling}
\label{sec:reward_design}
Existing RL approaches for CUDA generation \citep{baronio2025kevin,li2025cudaL1, anonymous2026mastering} use speedup over baselines as the reward signal. However, we observe that this approach suffers from outliers and bias toward easy kernels. This is because operators vary substantially in optimization difficulty, making raw speedup an unreliable proxy for code quality. To address this, we propose a normalized, robust reward scheme to remedy this and jointly optimize correctness and execution latency. 

We assign a reward score $r \in \{-1, 1, 2, 3\}$ based on correctness and performance:
\begin{equation}
r = \begin{cases} 
-1 & \text{if correctness check fails} \\ 
3 & \text{if } b(t, t_{\text{eager}}) \land b(t, t_{\text{compile}}) \\ 
2 & \text{if } b(t, t_{\text{eager}}) \\ 
1 & \text{otherwise} 
\end{cases}
\end{equation}
where $t$ is the generated kernel's runtime, $t_{\text{native}}$ and $t_{\text{compile}}$ are the runtimes of PyTorch's eager implementation and \texttt{torch.compile} version respectively, and $b(t, t_0) = \mathbb{I}\left[(t_0 - t) / t_0 > 5\%\right]$ indicates a significant speedup over baseline $t_0$. We validate this design in \cref{sec:ablation_reward} that it significantly outperforms the commonly used speedup reward.


\textbf{Efforts to Avoid Reward Hacking} We note that previous work \citet{lange2025sakanacuda} suffer from the hacking issue \citep{main_horse2025cuda}. Specifically, our system enforces the following constraints to avoid reward hacking: 
1) the provided Python scripts for correctness verification and performance profiling are protected via file permission controls, preventing the agent from modifying or interfering with the evaluation logic; 
2) to avoid trivial fallbacks, we enforce execution-time constraints using context managers that explicitly forbid invoking fallback implementations from \texttt{torch.nn.functional}, ensuring that performance gains originate solely from the generated CUDA kernels; 
3) for each problem, we validate kernel outputs against five randomly sampled inputs, strictly following the KernelBench evaluation protocol to ensure functional correctness; 
4) the profiling pipeline is carefully engineered with proper device synchronization, warm-up iterations, and repeated measurements with averaging, substantially reducing measurement noise and metric fluctuation; and 
5) the agent is not provided with any web search or external information retrieval tools, ensuring that all solutions are derived purely from local execution environment. 

\subsection{Algorithmic Improvements for Stable RL Training}
\label{sec:rl_training}
We observe that our initial RL trial could only train stably for 17 steps before the model's performance collapsed. We identified the root cause of this instability and propose warming up both the actor and critic models to adapt to the model's distribution. 
With this modification, our RL trial can stably train for 200 steps with consistent reward growth.

\textbf{The Root Cause of Training Instability} stems from a severe domain distribution mismatch, where the base model’s learned prior deviates significantly from the data distribution required for CUDA kernel coding. This is because the CUDA coding data accounts for less than 0.01\% of pretraining data~\citep{li2023starcoder, kocetkov2022stack}. This distribution gap results in numerous sampled low-probability and CUDA kernel code tokens. Furthermore, when training and inference engines use different numerical precisions (e.g., BF16 vs FP16), those low-probability tokens result in large importance sampling ratio variance because small numerical errors in computing token probablities $\pi_{\theta}(y_t)$ near the precision floor (e.g., $\pi_\theta(a_t \mid s_t) \approx 10^{-9}$) cause the importance ratio $\rho_t(\theta) = \frac{\pi_\theta(a_t \mid s_t)}{\pi_{\theta_{\text{old}}}(a_t \mid s_t)}$ to fluctuate wildly or explode (similar to the discussion in \citet{liu-li-2025-rl-collapse}). 



To achieve stable reinforcement learning, we propose a simple yet effective warm-up strategy: initializing both the actor and critic models using agent trajectories generated by the base model after single-turn RL, as illustrated in \cref{fig:training-pipeline}.


\textbf{Single-Turn Warm-up}~We first perform single-turn RL on base model to enhance its capability in CUDA kernel generation. We use PPO for optimization, where the base model serves as the policy and value network.

\textbf{Actor Initialization}~We then adopt a Rejection Fine-Tuning (RFT) stage on agent trajectories to initialize the actor model $\pi_{\theta}$. The resulting model from single-turn RL is used to collect CUDA agent trajectories by running the agent in the agent loop from \cref{sec:agent_loop}. Next, we apply RFT on the collected CUDA agent trajectories. Rejection sampling is performed to retain only high-quality rollouts according to the following rubrics:
(1) \textbf{Outcome filtering:} we only keep trajectories that achieve a positive reward ($R > 0$).
(2) \textbf{Pattern filtering:} we discard trajectories that exhibit inefficient or invalid behaviors, such as redundant multi-turn loops or hallucinations that violate the predefined tool-call schema.

The filtered trajectories are then used to optimize the actor via standard supervised fine-tuning with the following objective:
\begin{equation}
\mathcal{L}_{\text{RFT}}(\theta) = -\mathbb{E}_{\tau \sim \mathcal{D'}} \left[ \sum_{t=1}^{T} \log \pi_\theta(a_t \mid s_t, a_{<t}) \right],
\end{equation}
where $\tau = (s_0, s_1, \ldots, s_{T-1})$ denotes a filtered CUDA agent trajectory, $\pi_\theta$ is the policy parameterized by $\theta$, and $\mathcal{D'}$ represents the dataset after rejection sampling.

\begin{table*}[t]
\caption{\textbf{Main Results on KernelBench.} We report Pass Rate, Faster Rate (percentage of kernels faster than baseline), and Geometric Mean Speed-up. Metrics are reported relative to both PyTorch Eager and PyTorch Compile baselines. Overall metrics are weighted by the number of problems in each level (Level 1: 100, Level 2: 100, Level 3: 50). \textbf{Bold} indicates the best performance. } 
\label{tab:main_results}
\vskip 0.15in
\begin{center}
\resizebox{0.8\textwidth}{!}{
\begin{tabular}{llccccc}
\toprule
& & & \multicolumn{2}{c}{Faster Rate ($\uparrow$)} & \multicolumn{2}{c}{Speed-up (Geomean, $\times$)} \\
\cmidrule(lr){4-5} \cmidrule(lr){6-7}
Subset & Model & Pass Rate & vs. Eager & vs. Compile & vs. Eager & vs. Compile \\
\midrule
\multirow{3}{*}{Overall$^\dagger$}
& Seed1.6 (base model) & 74.0\% & 43.6\% & 27.2\% & 0.95$\times$ & 0.69$\times$ \\
& GLM 4.6 & 75.6\% & 44.8\% & 19.2\% & 0.78$\times$ & 0.57$\times$ \\
& Kimi K2 & 66.8\% & 40.8\% & 22.8\% & 0.93$\times$ & 0.66$\times$ \\
& Gemini 3 Pro & 91.2\% & 87.6\% & 69.6\% & 1.92$\times$ & 1.42$\times$ \\
& Claude Opus 4.5 & 95.2\% & 90.4\% & 66.4\% & 1.99$\times$ & 1.46$\times$ \\
& \ours (ours) & \textbf{98.8\%} & \textbf{98.4\%} & \textbf{96.8\%} & 
\textbf{2.60$\times$} & \textbf{2.11$\times$} \\
\midrule
\multirow{3}{*}{Level 1}
& Seed1.6 (base model) & 90.0\% & 63.0\% & 51.0\% & 1.65$\times$ & 1.25$\times$ \\
& GLM 4.6 & 86.0\% & 57.0\% & 32.0\% & 0.99$\times$ & 0.73$\times$ \\
& Kimi K2 & 85.0\% & 56.0\% & 39.0\% & 1.43$\times$ & 1.00$\times$ \\
& Gemini 3 Pro & 95.0\% & 90.0\% & 72.0\% & 1.99$\times$ & 1.51$\times$ \\
& Claude Opus 4.5 & 96.0\% & 88.0\% & 72.0\% & 2.03$\times$ & 1.54$\times$ \\
& \ours (ours) & \textbf{100.0\%} & \textbf{99.0\%} & \textbf{97.0\%} & \textbf{2.48$\times$} & \textbf{1.87$\times$} \\
\midrule
\multirow{3}{*}{Level 2}
& Seed1.6 (base model) & 74.0\% & 40.0\% & 16.0\% & 0.68$\times$ & 0.50$\times$ \\
& GLM 4.6 & 76.0\% & 43.0\% & 11.0\% & 0.60$\times$ & 0.42$\times$ \\
& Kimi K2 & 65.0\% & 40.0\% & 15.0\% & 0.93$\times$ & 0.65$\times$ \\
& Gemini 3 Pro & 93.0\% & 91.0\% & 76.0\% & 2.03$\times$ & 1.46$\times$ \\
& Claude Opus 4.5 & 98.0\% & 97.0\% & 69.0\% & 2.24$\times$ & 1.60$\times$ \\
& \ours (ours) & \textbf{100.0\%} & \textbf{100.0\%} & \textbf{100.0\%} & \textbf{3.27$\times$} & \textbf{2.80$\times$} \\
\midrule
\multirow{3}{*}{Level 3}
& Seed1.6 (base model) & 42.0\% & 12.0\% & 2.0\% & 0.60$\times$ & 0.40$\times$ \\
& GLM 4.6 & 54.0\% & 24.0\% & 10.0\% & 0.83$\times$ & 0.62$\times$ \\
& Kimi K2 & 34.0\% & 12.0\% & 6.0\% & 0.40$\times$ & 0.29$\times$ \\
& Gemini 3 Pro & 80.0\% & 76.0\% & 52.0\% & 1.58$\times$ & 1.17$\times$ \\
& Claude Opus 4.5 & 88.0\% & 82.0\% & 50.0\% & 1.52$\times$ & 1.10$\times$ \\
& \ours (ours) & \textbf{94.0\%} & \textbf{94.0\%} & \textbf{90.0\%} & \textbf{1.80$\times$} & \textbf{1.52$\times$} \\
\bottomrule
\end{tabular}
}
\end{center}
\vskip -0.1in
\end{table*}

\textbf{Critic Initialization}~We perform Value Pretraining to initialize the critic, specifically, we utilize the sampled trajectory data comprising state sequences and their corresponding outcome rewards to pretrain the critic network. Let $\tau = (s_0, s_1, \ldots, s_{T-1})$ denote a trajectory where $s_t$ represents the state at token position $t$, and let $r$ denote the outcome reward assigned at the final token (i.e., $r_t = 0$ for $t < T-1$ and $r_{T-1} = r$). We compute target values using Generalized Advantage Estimation ~\citep{gae}:
\begin{equation}
    V_t^{\text{targ}} = V_\phi(s_t) + \hat{A}_t, \quad \text{where} \quad \hat{A}_t = \sum_{l=0}^{T-1-t} (\gamma\lambda)^l \delta_{t+l},
\end{equation}
and $\delta_t = r_t + \gamma V_\phi(s_{t+1}) - V_\phi(s_t)$ is the temporal difference error with $V_\phi(s_T) = 0$. We set $\gamma = 1$ and $\lambda = 0.95$ in our experiments. We optimize the critic parameters $\phi$ by minimizing the mean squared error:
\begin{equation}
    \mathcal{L}_{\text{VP}}(\phi) = \frac{1}{2} \mathbb{E}_{\tau \sim \mathcal{D}} \left[ \frac{1}{T} \sum_{t=0}^{T-1} \left( V_\phi(s_t) - V_t^{\text{targ}} \right)^2 \right],
\end{equation}
where $\mathcal{D}$ denotes the collection of agent trajectories. We ablate the proposed initialization stage in \cref{sec:ablation_stages}.

\textbf{RL Algorithm}~\label{sec:PPO}
We employ PPO~\citep{schulman2017proximal} to optimize the actor model $\pi_{\theta}$. 
Let $\pi_{\theta}$ denote the policy to be optimized, and $\pi_{\theta_{\text{old}}}$ denote the policy used for trajectory sampling. We maximize the expected return using the clipped surrogate objective:
\begin{equation}
\begin{aligned}
    \mathcal{L}^{\text{CLIP}}(\theta) = \mathbb{E}_{\tau \sim \mathcal{D}} &\bigg[ \frac{1}{T} \sum_{t=0}^{T-1} \min \big(  \rho_t(\theta)\hat{A}_t, \\
    & \text{clip}(\rho_t(\theta), 1-\epsilon_{\text{lower}}, 1+\epsilon_{\text{higher}})\hat{A}_t \big) \bigg]
\end{aligned}
\end{equation}

where $\rho_t(\theta) = \frac{\pi_\theta(a_t \mid s_t)}{\pi_{\theta_{\text{old}}}(a_t \mid s_t)}$ is the importance sampling ratio between the current and old policies, $a_t$ denotes the action (i.e., token) taken at position $t$, $\epsilon_{\text{lower}}=0.2, \epsilon_{\text{higher}}=0.28$ following \citet{yu2025dapo}.

\section{Experiments}
\begin{table*}[t]
\caption{\textbf{Ablation Study.} Comparison between the full model and leave-one-out variants \textbf{under agent loop evaluation}. We analyze the contributions of (1) the agent loop, (2) robust reward design, (3) RFT, and (4) Value Pretraining. For variants without RFT or Value Pretraining, we report results from the final validation step before training collapse. 
}
\label{tab:ablation_res}
\vskip -0.15in
\begin{center}
\resizebox{0.85\textwidth}{!}{
\begin{tabular}{llccccc}
\toprule
& & & \multicolumn{2}{c}{Faster Rate ($\uparrow$)} & \multicolumn{2}{c}{Speed-up (Geomean, $\times$)} \\
\cmidrule(lr){4-5} \cmidrule(lr){6-7}
Subset & Model & Pass Rate & vs. Eager & vs. Compile & vs. Eager & vs. Compile \\
\midrule
\multirow{3}{*}{Overall$^\dagger$}
& w/o Agent Loop & 77.1\% & 43.5\% & 14.1\% & 0.89$\times$ & 0.69$\times$ \\
& w/o Robust Reward & 96.8\% & 90.4\% & 60.4\% & 1.70$\times$ & 1.25$\times$ \\
& w/o RFT & 95.6\% & 82.0\% & 49.8\% & 1.56$\times$ & 1.05$\times$ \\
& w/o Value Pretraining & 98.6\% & 85.0\% & 50.9\% & 1.49$\times$ & 1.00$\times$ \\
& \ours (ours) & \textbf{98.8\%} & \textbf{98.4\%} & \textbf{96.8\%} & 
\textbf{2.60$\times$} & \textbf{2.11$\times$} \\
\bottomrule
\end{tabular}
}
\end{center}
\vskip -0.1in
\end{table*}

\subsection{Experiment Settings}
\label{sec:experiments_setting}

\textbf{Training Settings for RL.} 
We leverage Seed1.6 \cite{seed16thinking} as the base model, a Mixture-of-Experts (MoE) model with 23B active and 230B total parameters.
We set the global batch size to 1024, matching the mini-batch size for online PPO updates. Learning rate for actor and critic is $3\times{10^{-6}}$ and $6\times{10}^{-6}$. These hyper-parameters are shared between single-turn RL and agentic RL. Context window length is 32768 for single-turn RL and 131072 for agentic RL. We impose a maximum of 150 agent turns during training rollouts, which is relaxed to 200 agent turns during evaluation. The model is trained over 150 training steps.

\textbf{Sandbox Environment} To ensure accurate speedup measurement for reliable reward signals and efficient GPU utilization, we design a CPU–GPU resource–decoupled sandbox architecture. A Docker-based terminal sandbox handles CPU-centric tasks (e.g., kernel compilation), while the agent leverages pre-defined \ours skill scripts to dispatch verification and profiling jobs to a dedicated GPU sandbox pool (128 NVIDIA H20 GPUs). This process-level isolation and exclusive resource allocation eliminate inter-process interference, ensuring stable latency measurements and guaranteed HBM capacity.

\textbf{Benchmark.} We conduct our evaluations on KernelBench~\cite{ouyang2025kernelbench}, utilizing the Level 1 to Level 3 subsets which comprise a total of 250 distinct operator tasks. To ensure fair comparison, we adapt these tasks from their original single-file format into our multi-file development environment (Section~\ref{sec:agent_loop}).

\textbf{Baseline Models.} We compare our approach against frontier proprietary coding models: \textbf{Claude Opus 4.5}~\citep{claudeopus45_systemcard_2025} and \textbf{Gemini 3 Pro} ~\citep{gemini3pro_modelcard_2025}, as well as two powerful open-source coding models: \textbf{GLM 4.6}~\citep{zai2025glm46} and \textbf{Kimi K2}~\citep{2025kimik2}. These models currently hold leading positions on major coding agent leaderboards and serve as strong baselines for general-purpose reasoning and coding capabilities. For fair comparison, baseline models are evaluated under the same agent loop mentioned in Section~\ref{sec:agent_loop}.

\textbf{Evaluation Metrics.} We assess performance using three key metrics: (1) \textbf{Pass Rate}, the percentage of tasks where the agent generates a kernel that successfully compiles and passes functional correctness checks; (2) \textbf{Faster Rate}, the percentage of tasks where the generated kernel is correct and achieves a faster execution time than the baseline (\texttt{eager} and \texttt{compile} modes); and (3) \textbf{Speed-up}, the geometric mean of the execution speed-up ratio relative to the baselines, computed exclusively for correct solutions.
To determine the final metrics for each task, we extract the best-performing solution from the agent's interaction trajectory, specifically the one that achieves the maximum speed-up relative to \texttt{torch.compile}. 

\subsection{Main Results}
\label{sec:main_results}

Table~\ref{tab:main_results} summarizes the performance of \ours against strong proprietary model baselines on KernelBench. Our analysis yields three primary insights regarding the efficacy of agentic RL for CUDA kernel optimization.

First, when compared to strong proprietary models, \ours delivers substantially stronger CUDA kernel coding performance. Although Claude Opus 4.5 and Gemini 3 Pro achieve respectable Pass Rates (91.2\%–95.2\%), their faster rates remain low at 66\%–69\%, indicating that general-purpose LLMs often produce naïve kernels that fail to outperform \texttt{torch.compile}. In contrast, \ours attains a 98.8\% Pass Rate and a 96.8\% faster rate, demonstrating that specialized RL training enables consistently correct and highly optimized CUDA implementations.


Second, compared to static \verb|torch.compile|, \ours prove that \emph{learned optimization policies can consistently outperform static compiler heuristics}, particularly in complex scenarios like operator fusion. This is most evident in Level 2 tasks (Operator Sequences), where \ours achieves a perfect 100\% faster rate and a massive 2.80$\times$ speed-up over \texttt{torch.compile}. Traditional compilers rely on predefined, rule-based patterns for kernel fusion which often struggle with non-trivial operator combinations. By contrast, \ours explores a much larger design space through its iterative agent loop, discovering hardware-specific memory access patterns and tiling strategies that remain inaccessible to static backends.


\subsection{Ablation Studies}
\label{sec:ablation}


\subsubsection{Impact of Skill-Integrated Agent Loop}
\label{sec:ablation_agent_loop}

To assess the critical role of the interactive environment in policy learning, we train two separate models under distinct protocols:
(1) \textbf{single-turn model}, trained using a standard code generation objective where the model predicts the final kernel and bindings in a single turn without execution feedback, serving as a warm-up stage in our training pipeline, and
(2) \textbf{\ours}, trained following the complete training pipeline, allowing the policy to see compilation errors and profiler feedback in multi-turn interactions.

Table~\ref{tab:ablation_res} highlights the limitations of single-turn code generation and underscores the necessity of our skill-integrated agent loop. Removing the agent loop causes a substantial drop in both correctness and optimization quality. More importantly, the resulting kernels are not only less optimized but often regress in performance. Being exposed to compilation errors, runtime failures, and profiler feedback, the agent can iteratively diagnose mistakes and refine transformations across turns.

\subsubsection{Impact of Reward Design}
\label{sec:ablation_reward}
To determine the optimal shaping of reward for kernel optimization, we evaluate two different reward formulations: (1) \textbf{speed-up reward}, a continuous reward signal defined as $r_{s} = t_{\text{compile}} / t_{\text{gen}}$ for correct solutions and $-1$ for incorrect ones; and (2) \textbf{our robust reward schedule} (\cref{sec:reward_design}), which assigns discrete values for achieving specific performance milestones .

As shown in Table~\ref{tab:ablation_res}, reward design has a pronounced effect on optimization outcomes. Replacing our robust reward schedule with a raw Speed-up Reward (w/o Robust Reward) yields comparable functional correctness, but substantially weaker optimization performance.

These results indicate that a normalized, milestone-based reward is better aligned with the goal of producing consistently faster kernels. By assigning credit to clear performance targets rather than directly regressing on noisy runtime ratios, the policy more reliably discovers transformations that translate into real speed-ups relative to both eager execution and compiler baselines.

\subsubsection{Impact of Multi-Stage Training}
\label{sec:ablation_stages}

As shown in Table~\ref{tab:ablation_res}, removing either RFT or Value Pretraining leads to substantial degradation in optimization performance, despite largely preserved pass rates.
More importantly, both ablations exhibit training instability and eventual collapse, motivating a closer analysis of the two stages below.

\begin{figure}[htbp]
  \centering
  \begin{subfigure}{0.24\columnwidth}
    \includegraphics[width=\linewidth]{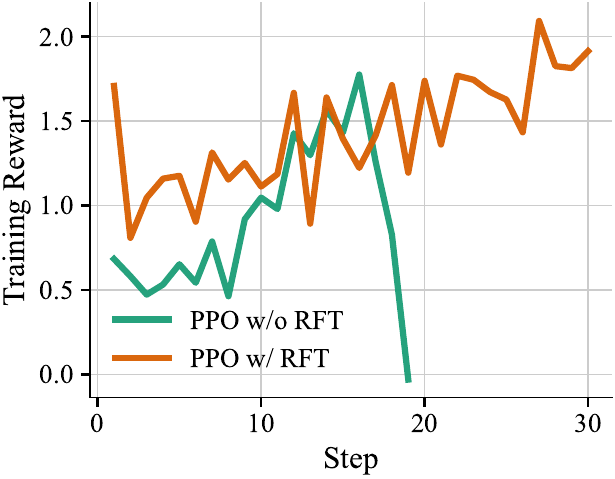}
    \caption{\textbf{Training Reward.}}
    \label{fig:rft_reward}
  \end{subfigure}
  \begin{subfigure}{0.24\columnwidth}
    \includegraphics[width=\linewidth]{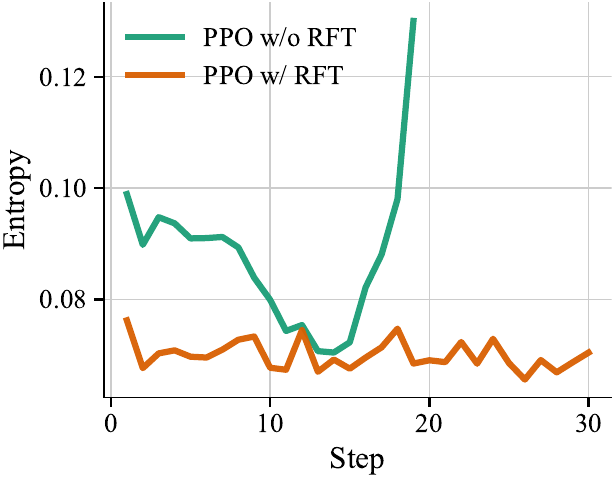}
    \caption{\textbf{Actor Entropy.}}
    \label{fig:rft_entropy}
  \end{subfigure}
\caption{\textbf{Ablation: RFT.} Removing RFT causes training reward to collapse. The concurrent increase in actor entropy suggests that the policy becomes increasingly diffuse and poorly structured.}

\end{figure}

\textbf{Rejection Sampling Fine-Tuning (RFT) provides a critical prior that prevents policy collapse.}
As shown in Figure~\ref{fig:rft_reward}, removing the RFT stage results in a rapid and catastrophic collapse of training rewards.
To diagnose the underlying cause, we examine the policy entropy in Figure~\ref{fig:rft_entropy}, which reveals a sharp increase coinciding with the reward collapse.
This entropy surge indicates that the policy distribution becomes increasingly diffuse, producing incoherent and poorly structured outputs. By initializing the policy with a strong behavioral prior, RFT constrains the entropy growth during reinforcement learning and keeps the optimization trajectory within a well-structured output distribution.


\begin{figure}[htbp]
  \centering
  \begin{subfigure}{0.24\columnwidth}
    \includegraphics[width=\linewidth]{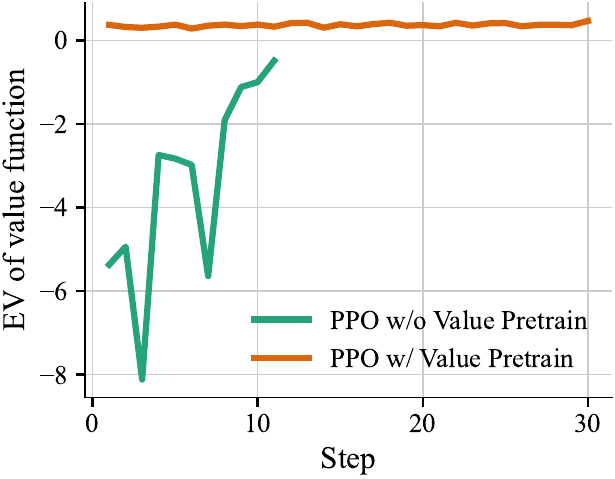}
    \caption{Explained Variation of Value Function.}
    \label{fig:vp_ev}
  \end{subfigure}
  \begin{subfigure}{0.24\columnwidth}
    \includegraphics[width=\linewidth]{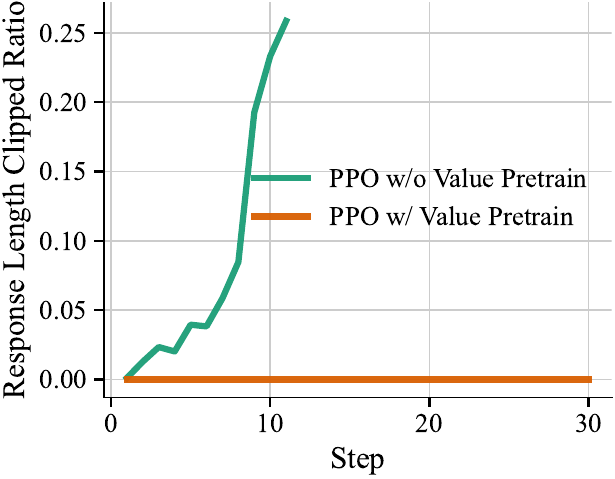}
    \caption{Response Length Clipped Ratio.}
    \label{fig:vp_clip}
  \end{subfigure}
  \caption{\textbf{Ablation: Value Pretraining.} Without Value Pretraining, the critic fails to learn a meaningful value function, as reflected by low explained variance. This leads to inefficient exploration, manifested as excessively long interaction trajectories.}

\end{figure}

\textbf{Value Pretraining is indispensable for providing reliable advantage estimates and preventing pathological search behaviors.} Without an initialized critic, the model fails to capture the value landscape of the multi-turn interaction states (Figure~\ref{fig:vp_ev}). This poor estimation leads to an explosion in trajectory length (Figure~\ref{fig:vp_clip}), as the uninitialized critic fails to penalize fruitless or redundant search paths. Value Pretraining ensures that the critic can immediately provide accurate feedback, guiding the agent toward efficient optimization paths and avoiding the computational instability of near-infinite interaction loops.

\section{Conclusion}

We introduced \ours, a large-scale agentic reinforcement learning system that endows large language models with the ability to generate and optimize CUDA kernels under realistic, execution-driven development workflows. By jointly scaling data synthesis, agent environments, and stability-oriented RL training, \ours moves LLMs beyond syntactic code generation toward hardware-aware performance optimization, achieving consistent gains over \texttt{torch.compile} and strong proprietary models on KernelBench. These results suggest a broader consequence: equipping foundation models with structured environments and reliable execution-based rewards can transform them from passive code generators into active systems optimizers, opening a path toward automating performance-critical software development in GPU computing.


\clearpage

\bibliographystyle{plainnat}
\bibliography{main}

\clearpage

\beginappendix

\section{Details of Collected Training Data}
\label{appd:data_details}

\begin{figure}[htbp]
  \centering
  \begin{minipage}{0.48\linewidth}
    \centering
    \includegraphics[height=8cm, keepaspectratio]{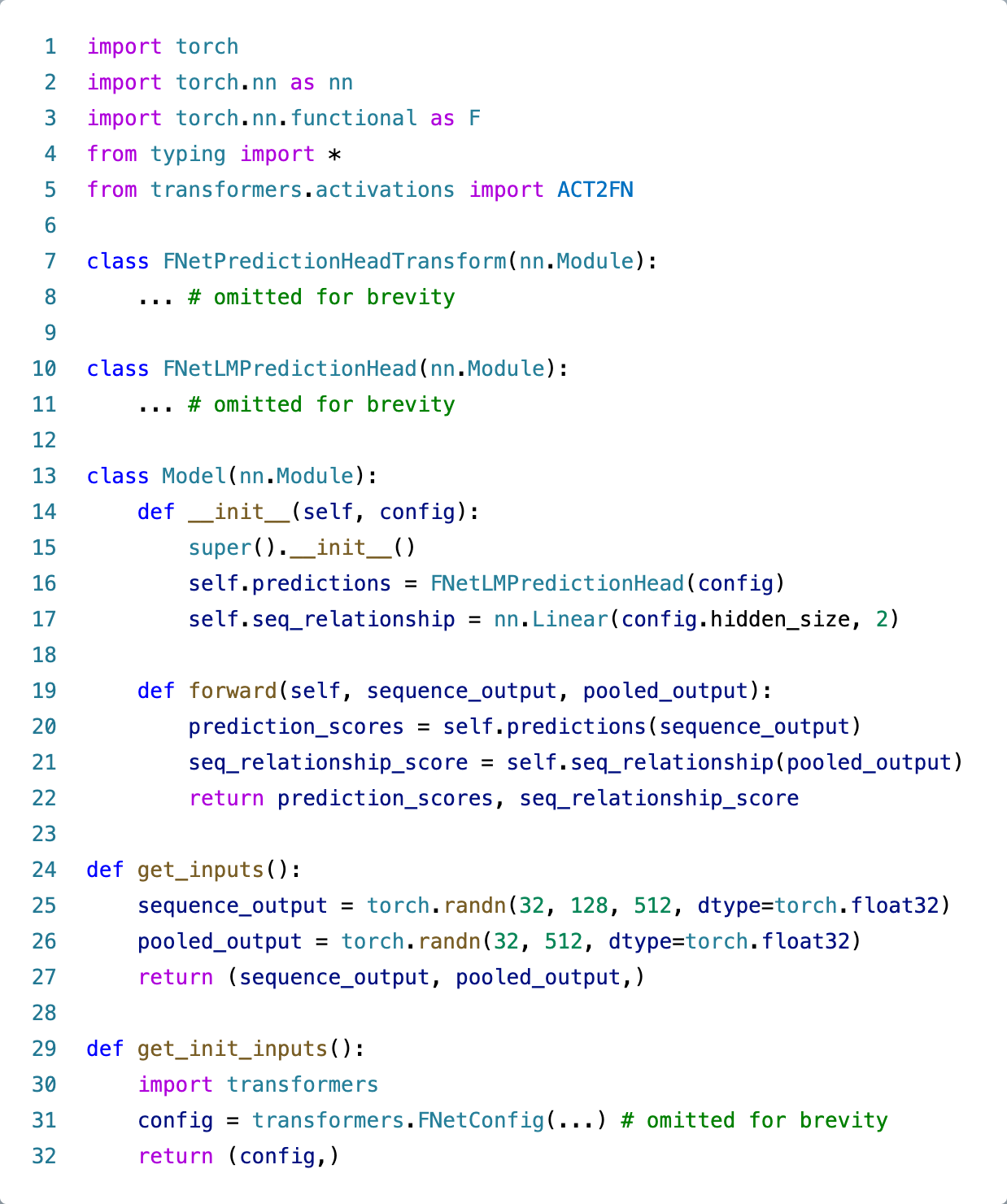}
    \caption*{(a) \texttt{transformers} operator class}
  \end{minipage}
  \hfill
  \begin{minipage}{0.48\linewidth}
    \centering
    \includegraphics[height=8cm, keepaspectratio]{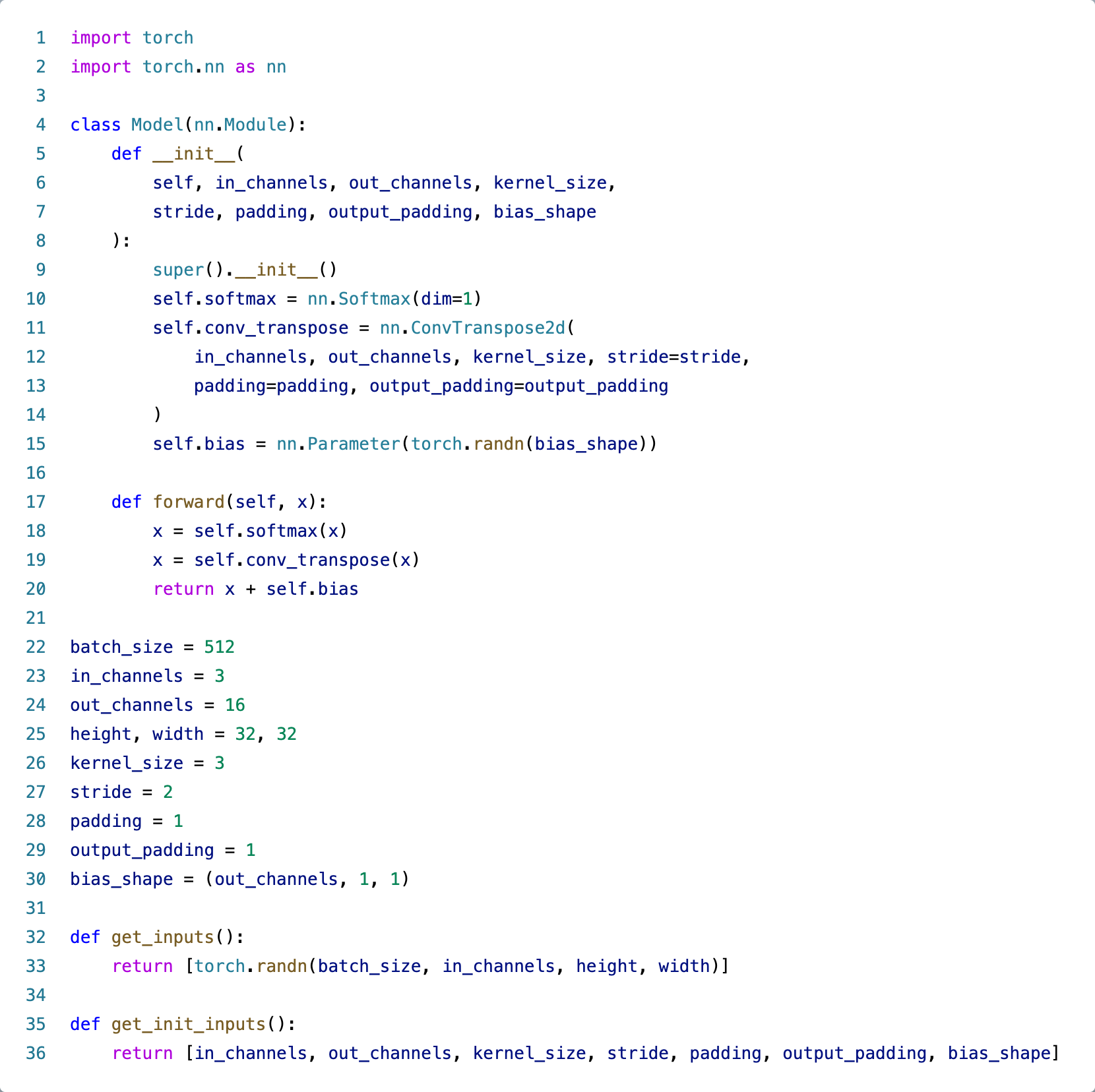}
    \caption*{(b) Combinatorial \texttt{torch} operator class}
  \end{minipage}
  \caption{Examples of operator classes in our training data.}
  \label{fig:operator_examples}
\end{figure}

\textbf{Operator Format.}
Each training sample is represented as a Python class implemented in PyTorch. Specifically, each operator is defined as a subclass of \texttt{torch.nn.Module}, with an \texttt{\_\_init\_\_} method for parameter initialization and a \texttt{forward} method specifying the computation.
In addition to the operator class, each sample includes two auxiliary functions: \texttt{get\_init\_inputs()}, which constructs the inputs required to instantiate the operator class, and \texttt{get\_inputs()}, which generates runtime inputs for the \texttt{forward} method. Together, the operator class and these auxiliary input-generation functions form a self-contained and executable training task. Figure~\ref{fig:operator_examples} shows representative examples of operator classes used in our training data.

\begin{figure}[htbp]
  \centering
  \includegraphics[width=0.6\linewidth]{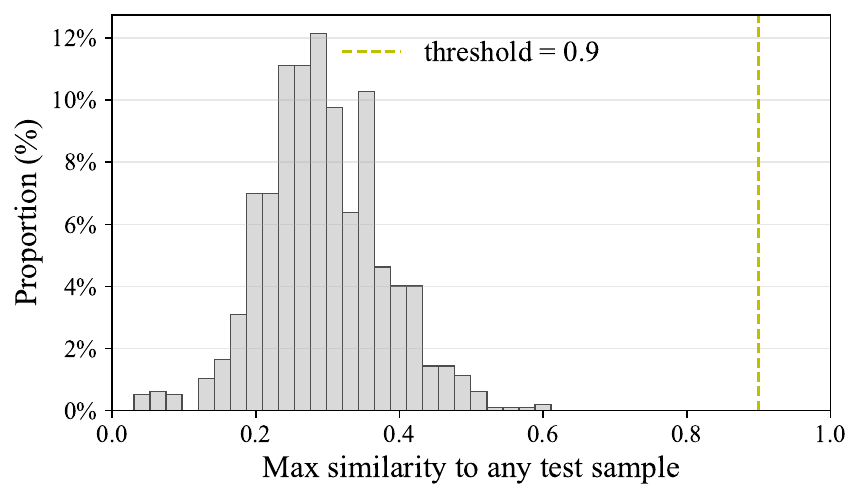}
  \caption{Distribution of the maximum AST similarity between each training sample and all evaluation samples.}
  \label{fig:max_sim_hist}
\end{figure}

\paragraph{Data Contamination Check.}
To prevent overlap between training data and evaluation benchmarks, we perform an explicit decontamination step using an off-the-shelf AST-based code similarity tool. Specifically, we extract the \texttt{Model} class from each program and compute pairwise structural similarity between training and evaluation samples using \texttt{PythonASTSimilarity}. A training sample is removed if its maximum similarity to any evaluation program exceeds 0.9. After this filtering, we confirm that no training sample exhibits high AST-level similarity with the evaluation set. Figure~\ref{fig:max_sim_hist} shows the distribution of the maximum AST similarity between each training sample and all evaluation samples. The majority of training samples exhibit low similarity scores, and no sample exceeds the decontamination threshold after filtering.

\begin{table}[htbp]
\centering
\caption{Composition of the final training dataset}
\label{tab:operator_distribution}
\begin{tabular}{l r}
\toprule
\textbf{Operator category} & \textbf{Proportion} \\
\midrule
\texttt{torch} operators $\times 1$ & 3.40\% \\
\texttt{torch} operators $\times 2$ & 83.77\% \\
\texttt{torch} operators $\times 3$ & 7.62\% \\
\texttt{torch} operators $\times 4$ & 2.80\% \\
\texttt{torch} operators $\times 5$ & 1.23\% \\
\texttt{transformers} operator & 1.18\% \\
\bottomrule
\end{tabular}
\end{table}

\paragraph{Final Dataset Composition.}
Table~\ref{tab:operator_distribution} summarizes the composition of the final training dataset after data synthesis and filtering. The majority of training samples are composite operator classes constructed by sequentially stacking between one and five operator classes from the \texttt{torch} library. We report the distribution of composite operators by composition depth, where a $k$-op composition denotes an operator class formed by composing $k$ primitive operator classes.
In addition, the dataset includes a set of operator classes directly taken from the \texttt{transformers} library, which are included as standalone operators and are not involved in compositional construction. This distribution reflects a deliberate balance between simple operators, moderately complex compositions, and higher-level \texttt{transformers} modules.
\section{Details of Agent Loop}
\label{appd:agent_loop_details}
\subsection{Full List of Provided Tools} 
We equip the agent with a structured toolset that abstracts common developer operations into callable interfaces. These tools define the action space of the agent and serve as the only mechanisms through which it can interact with the system and the external world.

The agent is provided with a suite of tools for controlled interaction with the local execution environment:

\textbf{Bash.}
Executes shell commands in a persistent session under strict safety constraints (e.g., command quoting rules and directory validation). It enables compilation, dependency management, and program execution.

\textbf{Read / Write.}
Provide read-only and write access to local files. Write operations are guarded by a read-before-write policy to prevent blind overwriting.

\textbf{Edit / MultiEdit.}
Support deterministic string-level code modifications. \textit{Edit} performs single replacements, while \textit{MultiEdit} enables multiple atomic edits within one file, ensuring consistency across dependent code changes.

\textbf{Glob.}
Performs fast file discovery using glob patterns (e.g., \texttt{**/*.py}), allowing the agent to navigate large codebases efficiently.

\textbf{Grep.}
A structured code search interface based on ripgrep, supporting regex search, file-type filtering, and contextual line retrieval.

\textbf{NotebookEdit.}
Enables structured modification of Jupyter notebook cells, allowing the agent to operate in mixed code–analysis environments.

Together, these tools allow the agent to perform typical software engineering operations, including code inspection, refactoring, compilation, and debugging.

\textbf{BashOutput.}
Streams incremental outputs from background shell processes, enabling the agent to monitor long-running jobs (e.g., training or compilation).

\textbf{KillBash.}
Terminates background shell sessions when jobs hang or resources must be reclaimed.

\subsection{Original SKILL.md Content} 

\begin{minted}[fontsize=\footnotesize]{text}
You are a PyTorch and CUDA expert. Accelerate the given PyTorch Model by creating a high-performance CUDA C++ extension, targeting the best possible performance with a minimum requirement of 5% faster than torch.compile baseline.

## 1. CRITICAL RESTRICTIONS

### STRICTLY FORBIDDEN
- **NO torch operators in C++**: NEVER use `torch::*` or `torch::nn::functional::*` in binding.cpp or .cu files
- **NO torch operations in model_new.py**: Only tensor creation and your custom ops allowed
- **NO third-party libraries**: Except cuBLAS (GEMM only) and cuDNN (Conv only)
- **NO modifications to utils/ directory**
- **NO modifications to binding.cpp or binding_registry.h**: These are fixed infrastructure

### ALLOWED ONLY
- **C++**: Raw CUDA kernels (for custom ops), cuBLAS (for GEMM), cuDNN (MANDATORY for Conv/ConvTranspose)
- **Python**: torch.tensor creation, custom extension ops, tensor properties (.shape, .device)
- **Memory**: torch::empty_like for allocation only
- **Focus**: Implement kernels in `kernels/` directory only

## 2. WORKSPACE STRUCTURE

```
.
binding_registry.h    # Do NOT modify - registration system
binding.cpp           # Do NOT modify - main module binding
kernels/              # YOUR WORK: Implement all kernels here
utils/                # DO NOT modify - Compilation, verification and profiling tools 
model.py              # DO NOT modify - Original PyTorch model
model_new.py          # YOUR WORK: Your optimized model using custom ops.
```

### File Types and Usage
- **`.cu` files**: CUDA kernels with `__global__` functions (custom implementations)
- **`.cpp` files**: cuDNN/cuBLAS API calls (NO custom kernels)
- **`_binding.cpp` files**: PyTorch tensor handling and Python bindings

## 3. UNIFIED WORKFLOW

### Step 1: Implementation

Create paired files in `kernels/`:

**kernels/my_kernel.cu** (Pure CUDA implementation):
```cuda
#include <cuda_runtime.h>

// Template kernel for performance tuning
template<int BLOCK_SIZE, int TILE_SIZE>
__global__ void my_kernel_impl(float* output, const float* input, int size) {
    // Shared memory for tiling
    extern __shared__ float smem[];
    
    int tid = blockIdx.x * blockDim.x + threadIdx.x;
    int stride = blockDim.x * gridDim.x;
    
    // Grid-stride loop for large data
    for (int i = tid; i < size; i += stride) {
        // Kernel logic with optimizations
        output[i] = /* computation */;
    }
}

// C-interface launcher (no PyTorch dependencies)
extern "C" void my_kernel_launcher(
    float* output,
    const float* input,
    int size,
    int config,
    cudaStream_t stream
) {
    // Dynamic configuration selection
    int blocks = (size + 255) / 256;
    int shared_mem_size = 0;
    
    switch(config) {
        case 0: 
            shared_mem_size = 256 * sizeof(float);
            my_kernel_impl<256, 16><<<blocks, 256, shared_mem_size, stream>>>(
                output, input, size);
            break;
        case 1: 
            shared_mem_size = 128 * sizeof(float);
            my_kernel_impl<128, 32><<<blocks, 128, shared_mem_size, stream>>>(
                output, input, size);
            break;
        default:
            my_kernel_impl<256, 16><<<blocks, 256, 0, stream>>>(
                output, input, size);
    }
}
```

**kernels/my_kernel_binding.cpp** (PyTorch binding):
```cpp
// Use this two headers to replace torch/extension.h for faster compilation
#include <torch/types.h>
#include <torch/csrc/utils/pybind.h>

#include <cuda_runtime.h>
#include <c10/cuda/CUDAStream.h>
#include "../binding_registry.h"

// Declare launcher from .cu file
extern "C" void my_kernel_launcher(
    float* output,
    const float* input,
    int size,
    int config,
    cudaStream_t stream
);

// PyTorch wrapper with config parameter
torch::Tensor my_kernel_forward(torch::Tensor input, int config = 0) {
    // Input validation
    TORCH_CHECK(input.is_cuda(), "Input must be a CUDA tensor");
    TORCH_CHECK(input.is_contiguous(), "Input must be contiguous");
    TORCH_CHECK(input.dtype() == torch::kFloat32, "Input must be float32");
    
    auto output = torch::empty_like(input);
    
    // Get current CUDA stream (correct way)
    cudaStream_t stream = c10::cuda::getCurrentCUDAStream().stream();
    
    // Call CUDA launcher with config
    my_kernel_launcher(
        output.data_ptr<float>(),
        input.data_ptr<float>(),
        input.numel(),
        config,
        stream
    );
    
    return output;
}

// Registration function
void register_my_kernel(pybind11::module& m) {
    m.def("my_kernel_forward", &my_kernel_forward, 
          "My kernel forward",
          py::arg("input"),
          py::arg("config") = 0);
}

// Auto-register
REGISTER_BINDING(my_kernel, register_my_kernel);
```

#### Create model_new.py
```python
import torch
import torch.nn as nn
import cuda_extension

class ModelNew(nn.Module):
    def __init__(self, ...):  # MUST match Model signature exactly
        super().__init__()
        # Initialize parameters - preserve original structure for state_dict compatibility
        self.weight = nn.Parameter(torch.randn(...))
        self.bias = nn.Parameter(torch.zeros(...))
        
    def forward(self, x):
        # Use custom ops only - NO torch operations
        x = cuda_extension.my_kernel_forward(x, config=0)
        x = cuda_extension.gemm_forward(x, self.weight, self.bias)
        return x
```

### Step 2: Compile and Test
```bash
# Compile with architecture-specific optimizations
TORCH_CUDA_ARCH_LIST=9.0 bash utils/compile.sh

# Test in sandbox 
sudo python3 -m utils.verification
sudo python3 -m utils.profiling
```

### Step 3: Performance Optimization (IF NEEDED)

#### 3.1 Optimization Strategy (Priority Order)

**Priority 1: Algorithmic (>50% impact)**
- Kernel fusion - reduce memory traffic
- Shared memory tiling - improve data reuse
- Memory coalescing - consecutive access patterns

**Priority 2: Hardware Utilization (20-50% impact)**
- Vectorized loads (float2/float4)
- Warp-level primitives (__shfl_sync, __ballot_sync)
- Occupancy tuning (block size, register usage)

**Priority 3: Fine-tuning (<20% impact)**
- Instruction-level parallelism
- Mixed precision (FP16/TF32)
- Prefetching and double buffering

#### 3.2 Parameter Tuning (Last Resort)
Only when within 1.2x of target and algorithmic options exhausted:

```python
# tune_kernel.py - NO recompilation needed
import time, torch, cuda_extension

configs = [
    (0, "256_threads_16_tile"),
    (1, "128_threads_32_tile"),
    (2, "512_threads_8_tile")
]

# Test input
x = torch.randn(batch_size, features).cuda()

# Benchmark each config
best_config, best_time = 0, float('inf')
for config_id, name in configs:
    # Warmup
    for _ in range(10):
        cuda_extension.my_kernel_forward(x, config=config_id)
    torch.cuda.synchronize()
    
    # Measure
    start = time.perf_counter()
    for _ in range(100):
        cuda_extension.my_kernel_forward(x, config=config_id)
    torch.cuda.synchronize()
    elapsed = time.perf_counter() - start
    
    print(f"Config {name}: {elapsed:.4f}s")
    if elapsed < best_time:
        best_time, best_config = elapsed, config_id

print(f"Best: config {best_config} ({best_time:.4f}s)")
# Update model_new.py with best_config
```

### Step 4: Iteration Requirements

#### Correctness Failures
**MUST iterate until correctness passes - NO EXCEPTIONS**
1. Debug the specific failing kernel
2. Common issues to check:
   - Boundary conditions (tid < size)
   - Synchronization (__syncthreads placement)
   - Data types and precision
   - Memory alignment
3. Fix in kernels/*.cu and *_binding.cpp ONLY
4. Recompile and test

#### Performance Optimization
**GOAL: Achieve the best possible performance (the faster, the better!)**
**MINIMUM: Must be at least 5% faster than torch.compile baseline**

For each iteration:
1. **Document expectation**: "Fusion will eliminate 3 kernels, expect ~20% speedup"
2. **Apply optimization aggressively**: Don't revert to slow versions
3. **Debug if correctness fails**: Fix the optimized version
4. **Measure and analyze**: Understand why performance changed
5. **Continue optimizing**: Even if you meet the minimum, keep pushing for better performance

**Iteration strategy**:
- First 1-2 iterations: Achieve the minimum 5% improvement
- Next 3-5 iterations: Push for maximum possible speedup
- Continue until no further improvements possible or diminishing returns

**Remember**: The goal is the BEST possible performance, not just meeting the minimum!

### Step 5: Final Cleanup (MANDATORY BEFORE COMPLETION)

Before declaring the task complete, clean up the kernels/ directory to contain ONLY the final optimized version:

**Remove all intermediate attempts**:
```bash
# Remove version files, old attempts, test versions
rm kernels/*_v[0-9].cu kernels/*_old.cu kernels/*_test.cu kernels/*.bak

# Keep only the final optimized implementation
# Example final structure:
# kernels/
#   fused_kernel.cu           # Final implementation
#   fused_kernel_binding.cpp  # Final binding
```

## 4. TOOL SCRIPTS REFERENCE

### Verification and Profiling
```bash
# Use sudo to run sandbox utilities
sudo python3 -m utils.verification
sudo python3 -m utils.profiling
```

### Compilation
```bash
TORCH_CUDA_ARCH_LIST=9.0 bash utils/compile.sh
```


## 5. OPTIMIZATION CHECKLIST

### Essential Optimizations (Apply First)
- [ ] **Memory Coalescing**: Consecutive threads access consecutive addresses
- [ ] **Kernel Fusion**: Combine operations to reduce memory traffic
- [ ] **Shared Memory**: Cache frequently accessed data
- [ ] **Grid-Stride Loops**: Handle data larger than grid size
- [ ] **Boundary Checks**: Validate all array accesses (tid < size)

### Performance Optimizations (Apply as Needed)
- [ ] **Vectorized Memory**: Use float2/float4 for higher throughput
- [ ] **Warp Primitives**: __shfl_sync for inter-thread communication
- [ ] **Occupancy Tuning**: Balance block size and resource usage
- [ ] **Bank Conflict Avoidance**: Pad shared memory arrays
- [ ] **Loop Unrolling**: Increase instruction-level parallelism

### Advanced Optimizations (For Final Tuning)
- [ ] **Tensor Cores**: Use WMMA/MMA for eligible GEMM operations
- [ ] **Mixed Precision**: FP16/TF32 where appropriate
- [ ] **Persistent Kernels**: Keep data in registers across iterations
- [ ] **CUDA Graphs**: Reduce launch overhead
- [ ] **Double Buffering**: Overlap computation with memory transfers

### Correctness Checklist (Always Verify)
- [ ] **Thread Bounds**: Check tid < N before array access
- [ ] **Synchronization**: __syncthreads() before shared memory reuse
- [ ] **Data Types**: Ensure correct types and conversions
- [ ] **Memory Safety**: No out-of-bounds access
- [ ] **Numerical Stability**: Handle NaN/Inf, use stable algorithms

## 6. COMMON ISSUES AND SOLUTIONS

### Compilation Errors
| Error | Solution |
|-------|----------|
| undefined symbol | Check extern "C" declarations match |
| no kernel image | Verify TORCH_CUDA_ARCH_LIST matches GPU |

### Correctness Failures
| Issue | Debug Steps |
|-------|-------------|
| Wrong output values | 1. Check kernel math<br>2. Verify indexing<br>3. Test with simple inputs |
| NaN/Inf results | 1. Check division by zero<br>2. Verify numerical stability<br>3. Add bounds checking |
| Mismatched shapes | 1. Print tensor shapes<br>2. Check dimension calculations<br>3. Verify reduction logic |

### Performance Issues
| Symptom | Likely Cause | Solution |
|---------|--------------|----------|
| Slower than baseline | No fusion | Combine kernels |
| Low SM efficiency | Poor occupancy | Tune block size |
| Low memory throughput | Uncoalesced access | Restructure memory pattern |
| High kernel count | Missing fusion | Implement compound operations |

## 7. SUCCESS CRITERIA

**OPTIMIZATION GOALS:**
- **MINIMUM REQUIREMENT**: At least 5% faster than torch.compile (<= 0.95× baseline time)
- **TARGET**: Achieve the best possible performance - every microsecond counts!
- **Correctness**: Test must pass (atol=1e-2, rtol=1e-2)
- **Clean Final Code**: kernels/ directory contains ONLY final optimized version (no intermediate attempts)

**Performance metric clarification:**
- If torch.compile baseline = 1.0ms:
  - MINIMUM: Your implementation must be <= 0.95ms (5% faster)
  - GOAL: Push for <= 0.8ms or better (20%+ faster)
- The faster your implementation, the better the result
- Continue optimizing even after meeting the minimum requirement

## 8. KEY REMINDERS

1. **Keep .cu and _binding.cpp files separate** - Faster compilation
2. **Pass config parameters through bindings** - Enable runtime tuning without recompilation
3. **Focus modifications in kernels/ directory** - Never modify infrastructure files
4. **Be aggressive with optimizations** - Don't revert to slow versions when debugging
5. **Document performance expectations** - Before implementing, state expected gains
6. **Test with descriptive names** - Show which optimizations are applied
7. **Clean up before completion** - Remove ALL intermediate attempts from kernels/, keep ONLY final version

## Your Task

Optimize the PyTorch model in model.py.
\end{minted}
\section{Discussion of Concurrent Works}
\label{appd:concurrent_works}

In this appendix, we provide a more detailed discussion of concurrent and closely related works, clarifying differences in problem settings, evaluation protocols, and training assumptions that make direct comparison with our method either inappropriate or non-informative.

\paragraph{STARK.}
STARK adopts Claude Sonnet~4 as its base model and builds a structured multi-agent system consisting of specialized roles for planning, coding, and debugging. These agents follow a fixed execution program to explore a tree-structured search space. In contrast, our method employs a single agent that autonomously invokes tools, gathers feedback, and performs iterative kernel correction and optimization.
Although both methods are evaluated on KernelBench, the substantial difference between a fixed multi-agent pipeline and a single, autonomous agent makes direct comparison less meaningful. In addition, STARK does not explicitly specify the GPU hardware used for runtime evaluation, which further complicates precise performance comparison.

\paragraph{ReGraphT.}
ReGraphT focuses on a distinct research question: transferring the optimization capabilities of large language models to smaller models via retrieval-augmented reasoning graphs and Monte Carlo Graph Search. As its primary goal is model compression and capability distillation rather than maximizing absolute kernel generation performance, we do not include direct performance comparisons with ReGraphT.

\paragraph{EvoEngineer.}
EvoEngineer formulates kernel optimization as an evolutionary code editing process driven by LLM feedback. However, its evaluation is conducted on only 91 problems selected from the 250 tasks in KernelBench, rather than the full benchmark. This partial coverage introduces selection bias and limits the representativeness of the reported results, making direct comparison with our full-benchmark evaluation inappropriate.

\paragraph{CudaForge.}
CudaForge adopts a two-agent workflow built on OpenAI-o3, where a Judge agent leverages Nsight Compute profiling signals and hardware specifications to diagnose performance bottlenecks and provide targeted optimization feedback to a Coder agent that applies code edits. Similar to STARK, this approach relies on predefined agent roles and a fixed interaction protocol. In contrast, our method employs a single agent that autonomously decides when to invoke tools, interprets feedback, and iteratively revises kernels in a self-directed manner.

\paragraph{Kevin.}
Kevin introduces a multi-turn reinforcement learning framework that explicitly models the iterative CUDA development workflow. However, this approach partitions KernelBench into subsets for training and evaluation, effectively training on benchmark-derived tasks. As a result, the reported gains may partially reflect benchmark-specific adaptation rather than generalizable kernel generation capability, which limits the fairness of direct comparison with methods trained without access to KernelBench data.

\paragraph{CUDA-L1.}
CUDA-L1 directly constructs supervised fine-tuning data from KernelBench reference implementations and further applies reinforcement learning using execution-based rewards on the same benchmark tasks. This practice results in significant data leakage between training and evaluation, rendering the reported performance not directly comparable to approaches, such as ours, that strictly avoid using KernelBench for training.

\paragraph{ConCuR.}
ConCuR synthesizes CUDA kernels with reasoning traces generated by a Kevin-32B model and uses the resulting dataset to fine-tune KernelCoder. As previous mentioned, 
Kevin-32B is trained on a KernelBench subset, so the performance of KernelCoder do not reflect training from independently curated or real-world kernel optimization data.

\section{Case Study}
\label{appd:case_study}

we conduct an in-depth case study on \ours's optimization trajectories over Level~1 to Level~3 of KernelBench. For each level, we analyze a representative example and distill the key optimization applied by \ours.

\subsection{Common Optimization Patterns}

Across all three difficulty levels in KernelBench, we observe several recurring optimization patterns in the trajectories generated by \ours.

\paragraph{Algebraic Simplification and Operator Reduction.}
A dominant pattern is the use of algebraic reasoning to simplify high-level tensor expressions and reduce computational complexity. In Case~\ref{case:1}, an explicit diagonal matrix multiplication is reduced to row-wise scaling, eliminating an entire matrix construction and GEMM. In Case~\ref{case:2}, a matrix multiplication followed by reduction is transformed into a reduction over weights followed by a dot product. Such transformations replace generic, high-cost operators with simpler computations that are better aligned with the underlying mathematical structure.

\paragraph{Kernel Fusion.}
Another recurring pattern is kernel fusion for eliminating intermediate tensors. Sequences of logically related operations are merged into a single kernel to avoid intermediate tensor materialization and reduce kernel launch overhead. Examples include collapsing multiple arithmetic stages (summation, dot product, division, and scaling) into a single kernel in Case~\ref{case:2}, and fusing element-wise residual addition with activation functions in Case~\ref{case:3}. Fusion improves data locality and reduces global memory traffic.

\paragraph{Memory Access Optimization.}
Efficient memory usage emerges as a key theme across cases. The generated kernels emphasize coalesced global memory accesses, reduced memory footprint by avoiding large intermediate tensors, and judicious use of shared memory for intra-block reductions. Vectorized loads (e.g., \texttt{float4}) are employed when beneficial to maximize memory bandwidth utilization.

\paragraph{Hardware-Aware Optimization.}
For complex models, \ours demonstrates explicit awareness of GPU hardware capabilities and tailors the generated implementation accordingly. In Case~\ref{case:3}, the system enables TF32 computation for both matrix multiplications and convolution operations, allowing the execution to leverage Tensor Cores on modern GPUs. By selectively adopting lower-precision arithmetic where it is performance-critical yet numerically acceptable, \ours achieves substantial speedups without requiring manual intervention.

\paragraph{Library-Aware Optimization.}
In addition to custom kernel generation, \ours effectively leverages highly optimized vendor libraries when appropriate. In Case~\ref{case:3}, the system identifies opportunities to invoke fused cuDNN APIs, such as convolution with bias and activation, to replace multiple separate operators. By mapping high-level model semantics onto optimized library primitives, \ours reduces kernel launch overhead and benefits from mature, hardware-tuned implementations provided by cuDNN.

Overall, these common patterns illustrate how \ours systematically bridges algorithmic insight and low-level performance engineering to optimize diverse workloads ranging from simple linear algebra to full neural network building blocks.


\begin{figure}[htbp]
  \centering
  \includegraphics[width=0.8\linewidth]{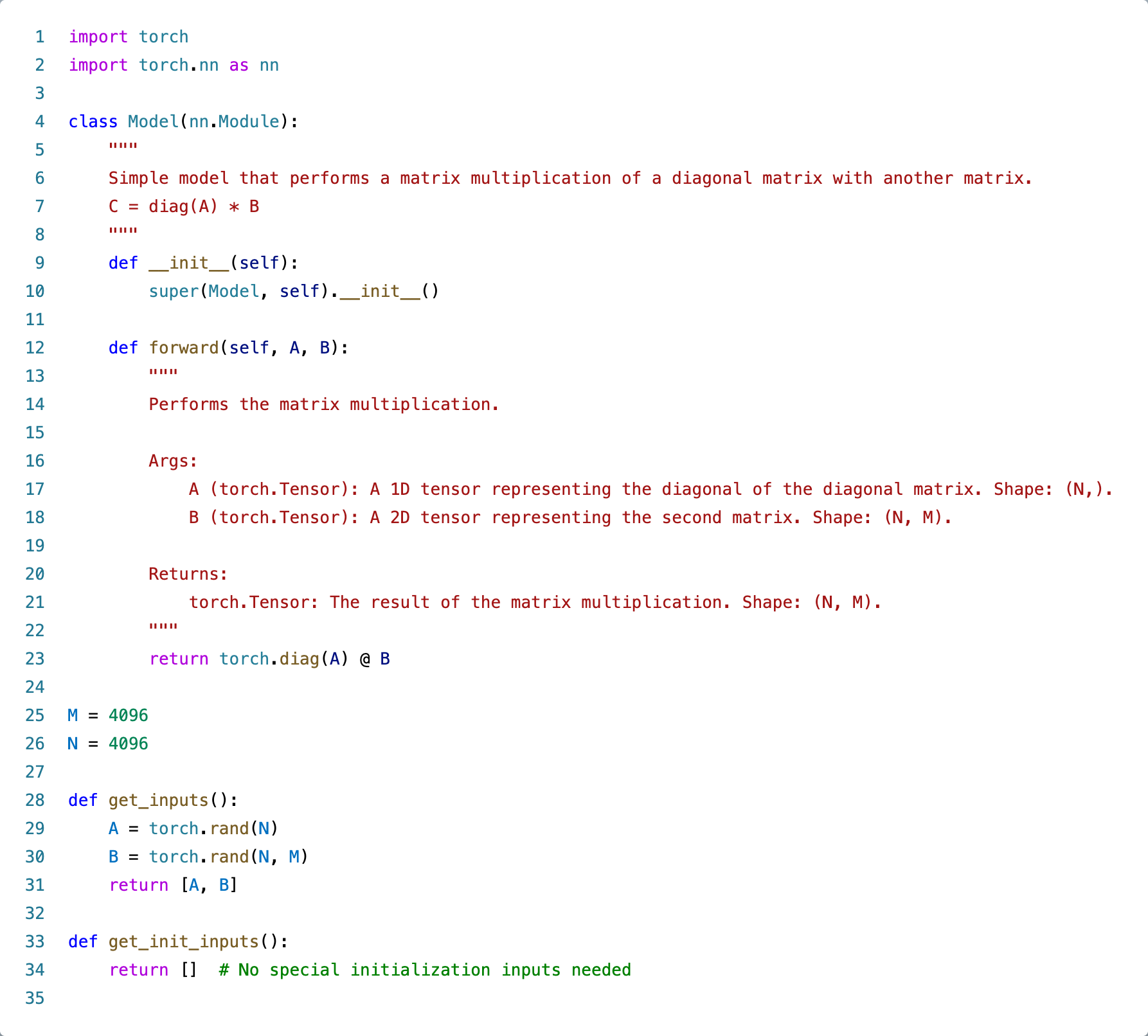}
  \caption{Reference operator for diagonal \texttt{matmul} (Case~\ref{case:1}).}
  \label{fig:level1_0_3_p1}
\end{figure}

\begin{figure}[htbp]
  \centering
  \includegraphics[width=0.8\linewidth]{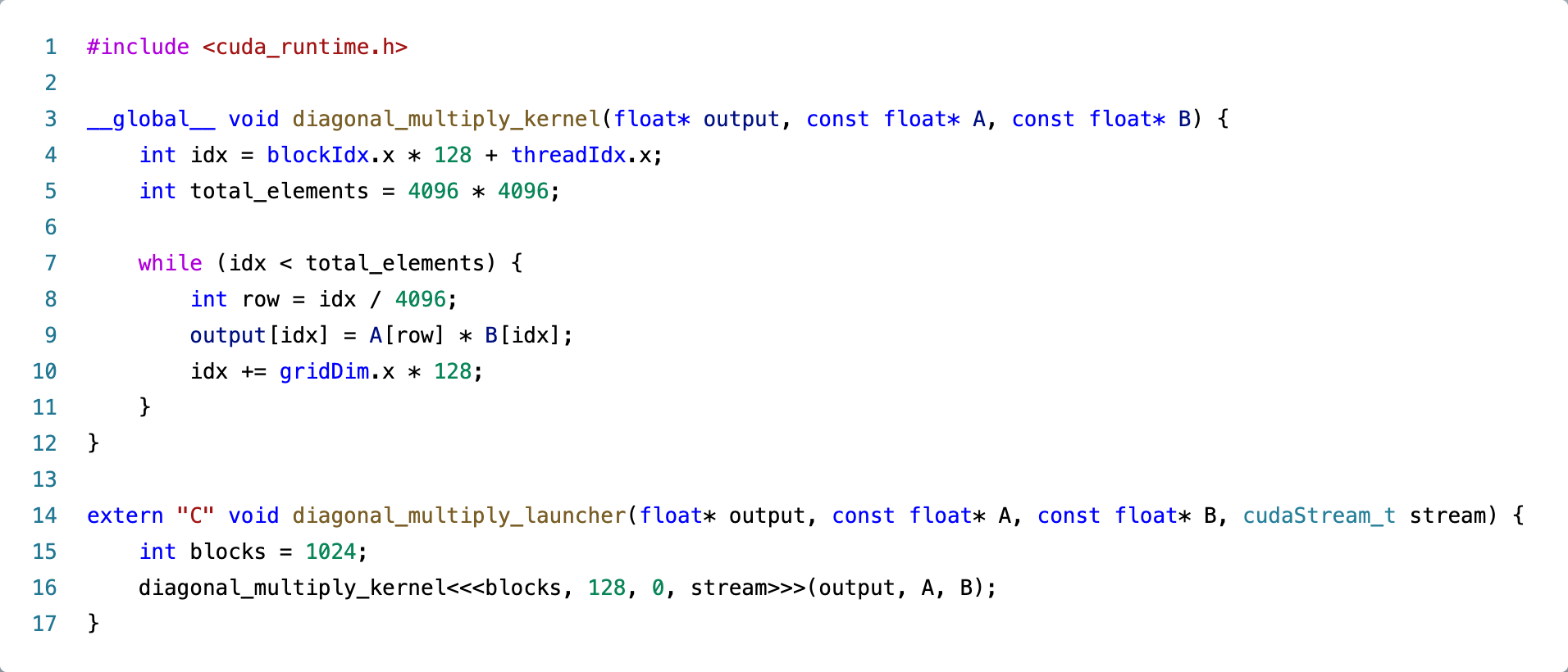}
  \caption{Diagonal \texttt{matmul} kernel implementation (Case~\ref{case:1}).}
  \label{fig:level1_0_3_p2}
\end{figure}


\begin{figure}[htbp]
  \centering
  \includegraphics[width=0.8\linewidth]{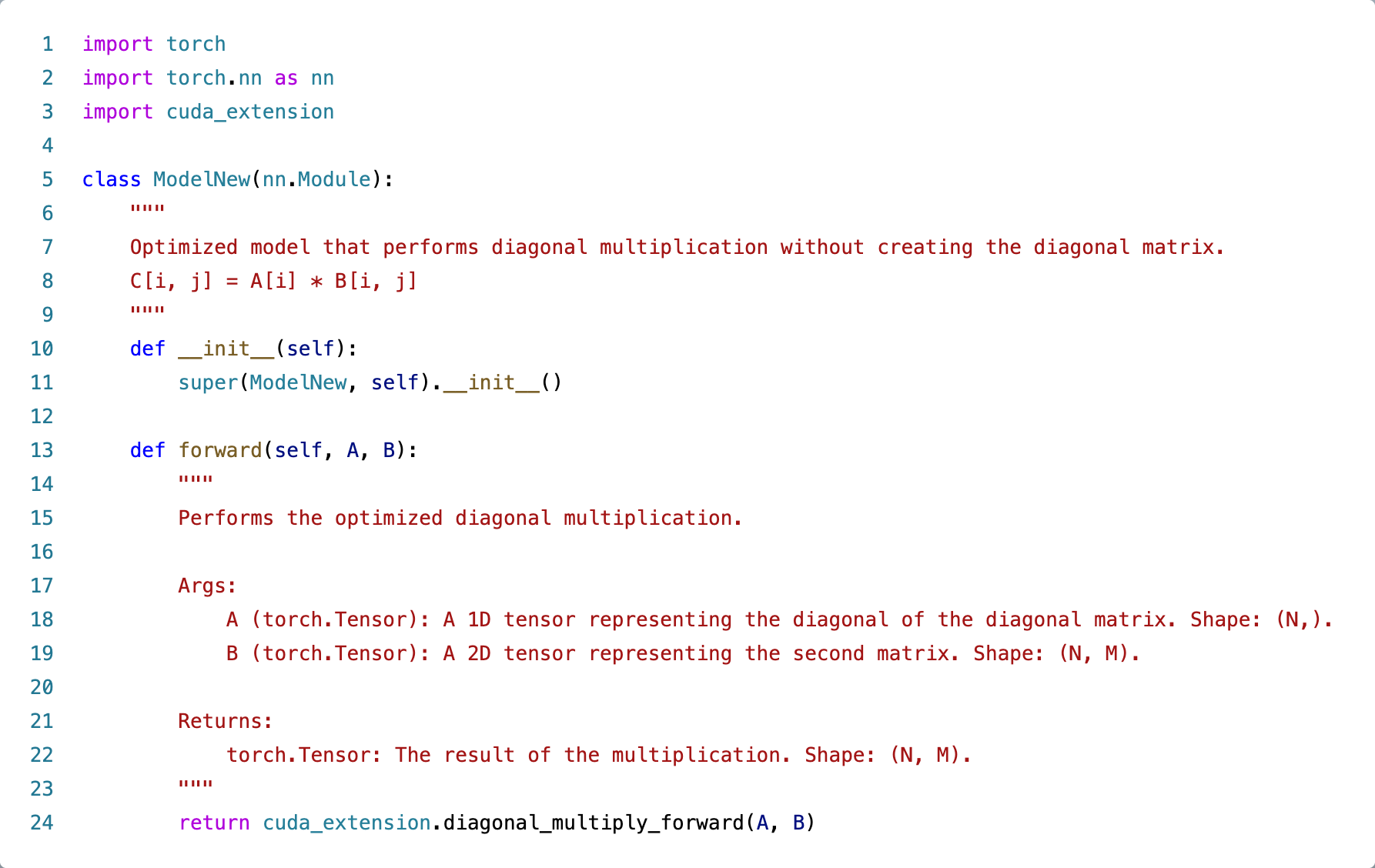}
  \caption{Custom operator for diagonal \texttt{matmul} (Case~\ref{case:1}).}
  \label{fig:level1_0_3_p4}
\end{figure}

\subsection{Example of Level~1: Model that performs matrix multiplication of a diagonal matrix with another matrix}
\label{case:1}

In this case, the reference model computes the product of a diagonal matrix constructed from a vector and a dense matrix. Although mathematically simple, this formulation incurs unnecessary computational and memory overhead by explicitly materializing the diagonal matrix and invoking a general matrix multiplication (GEMM). Figures~\cref{fig:level1_0_3_p1,fig:level1_0_3_p2,fig:level1_0_3_p4} illustrate the reference operator, the optimized CUDA kernel, and the resulting custom operator for this example. The Python/C++ binding is omitted for brevity. This custom operator achieves 73.31$\times$ speed-up versus Torch Compile.

\ours exploits the algebraic structure of diagonal matrix multiplication. Specifically, left-multiplying a matrix \( B \) by a diagonal matrix defined by vector \( A \) is equivalent to scaling each row \( i \) of \( B \) by the scalar \( A[i] \). This observation reduces the computation from a matrix--matrix multiplication to an element-wise broadcast multiplication, lowering the time complexity from \(O(N^2M)\) to \(O(NM)\).

Based on this simplification, \ours implements a custom CUDA kernel that directly performs the row-wise scaling without constructing the intermediate diagonal matrix. The kernel uses a grid-stride loop to efficiently cover all elements of the output matrix, and each thread independently multiplies one element of \( B \) by the corresponding diagonal entry from \( A \). This approach fuses the diagonal construction and matrix multiplication into a single kernel, significantly reducing kernel launch overhead and global memory traffic.

Overall, this case illustrates a common optimization pattern identified by \ours: recognizing implicit structure in high-level tensor expressions and replacing generic operators with specialized kernels that directly implement the underlying mathematical operation.


\begin{figure}[htbp]
  \centering
  \includegraphics[width=0.8\linewidth]{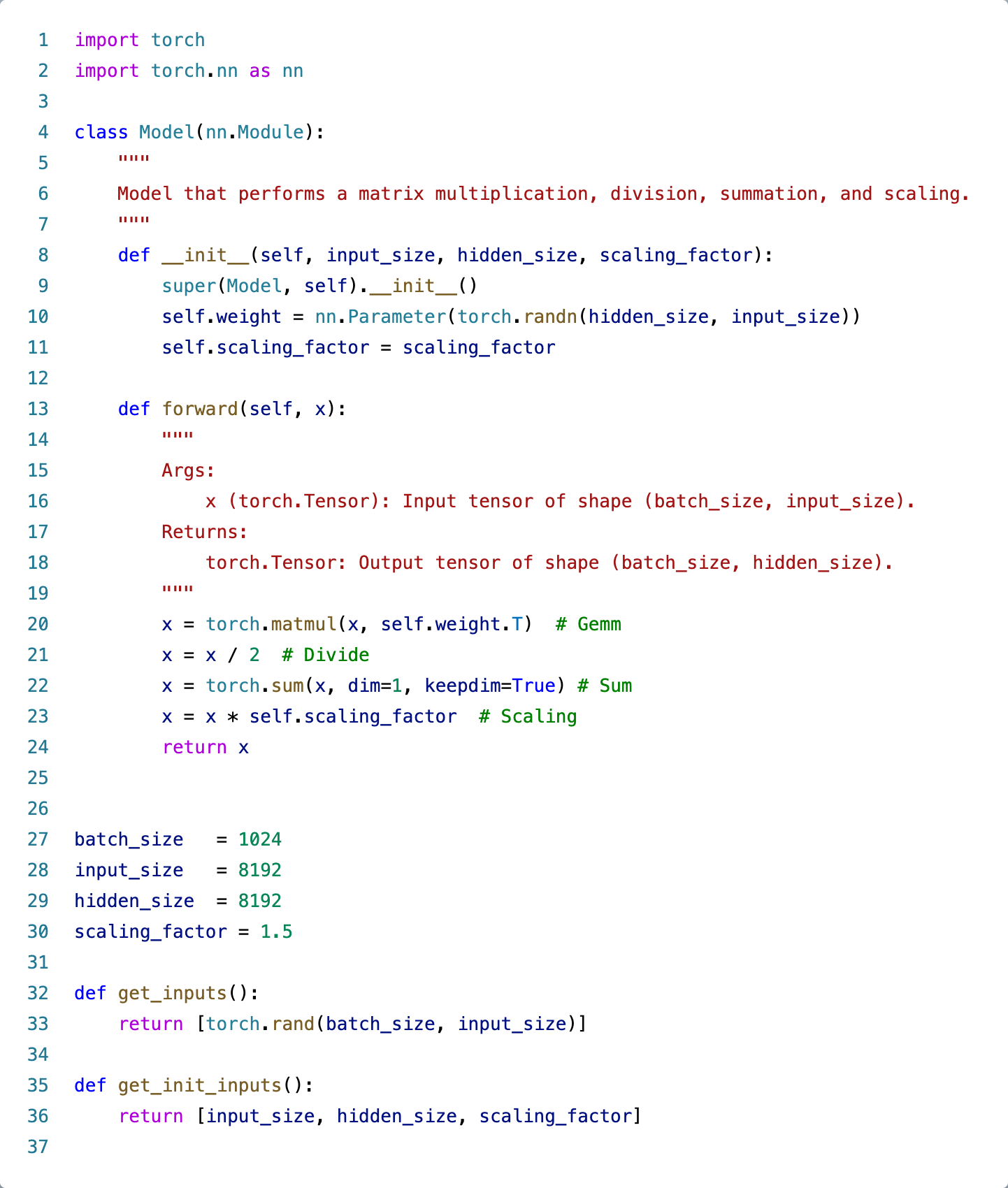}
  \caption{Reference operator for matrix multiplication, division, summation, and scaling (Case~\ref{case:2}).}
  \label{fig:level2_3_105_p1}
\end{figure}

\begin{figure}[htbp]
  \centering
  \includegraphics[width=0.75\linewidth]{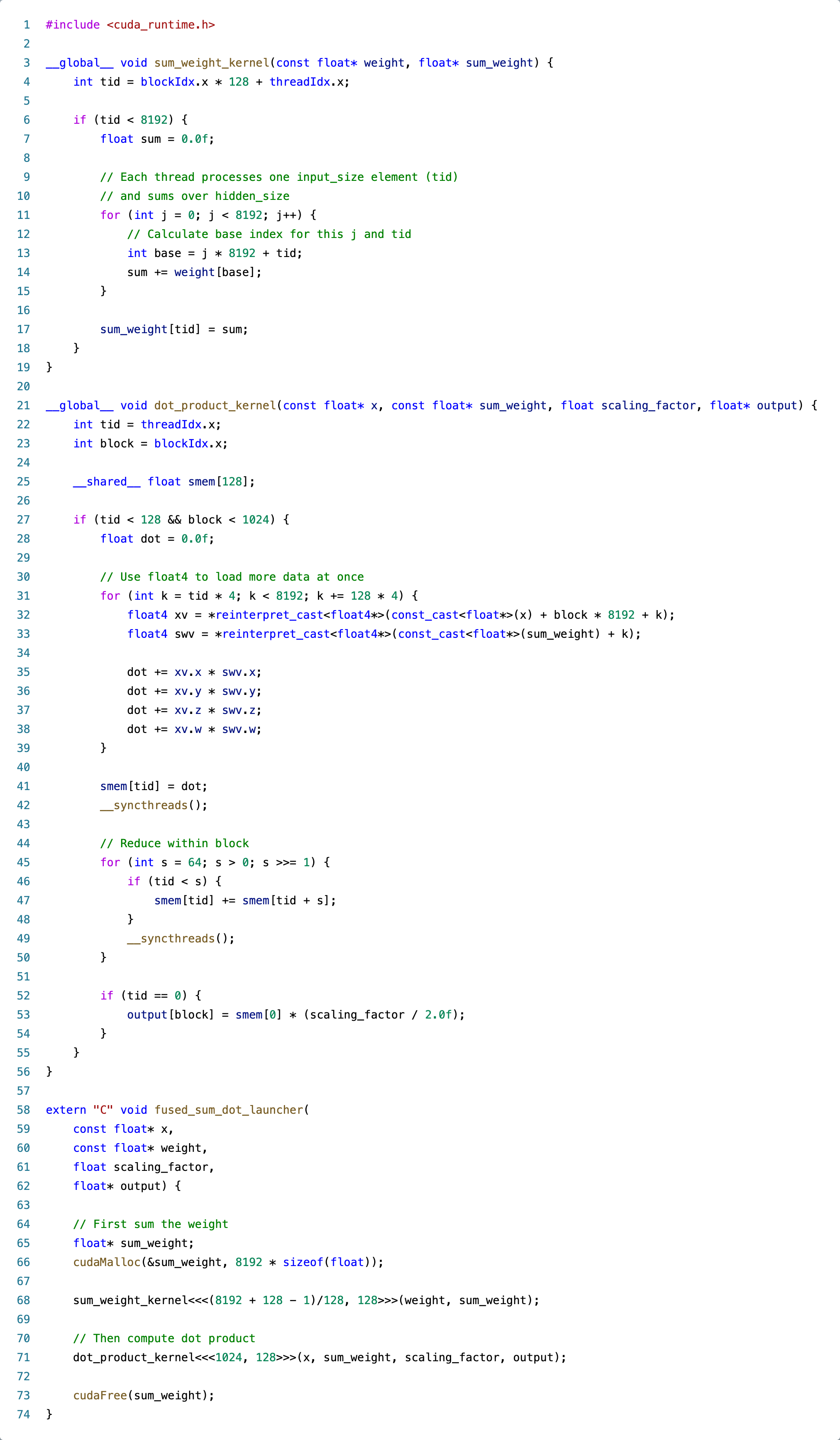}
  \caption{Fused \texttt{sum-then-dot-product} kernel implementation (Case~\ref{case:2}).}
  \label{fig:level2_3_105_p2}
\end{figure}


\begin{figure}[htbp]
  \centering
  \includegraphics[width=0.8\linewidth]{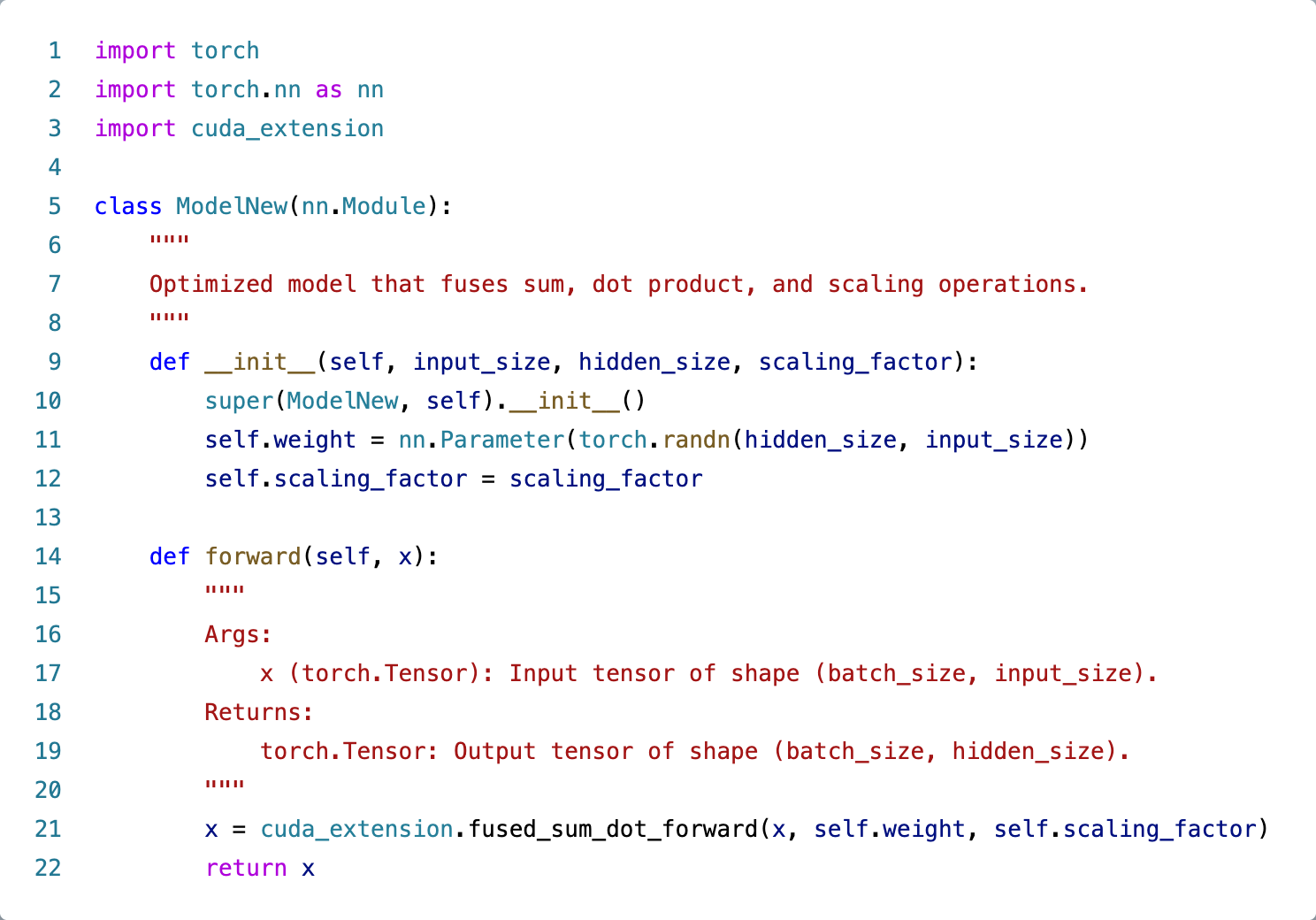}
  \caption{Custom operator for matrix multiplication, division, summation, and scaling (Case~\ref{case:2}).}
  \label{fig:level2_3_105_p4}
\end{figure}

\subsection{Example of Level~2: Model that performs matrix multiplication, division, summation, and scaling}
\label{case:2}

The second case involves a model composed of a sequence of operations, including matrix multiplication, division, summation, and scaling. Executed naively, this pipeline materializes large intermediate tensors and performs redundant computations. Figures~\cref{fig:level2_3_105_p1,fig:level2_3_105_p2,fig:level2_3_105_p4} show the reference operator, the fused CUDA kernel, and the final custom operator produced by \ours. The Python/C++ binding is omitted for brevity. This custom operator achieves 24.04$\times$ speed-up versus Torch Compile.

\ours begins with an algebraic rearrangement of the computation. By exploiting the linearity of summation and matrix multiplication, the operation can be rewritten as
\[
\sum_j \frac{x_i \cdot w_j^T}{2} = x_i \cdot \left(\sum_j w_j^T\right) / 2,
\]
transforming a matrix--matrix multiplication followed by a reduction into a column-wise reduction of the weight matrix followed by a dot product. This significantly reduces both the number of floating-point operations and memory accesses.

\ours implement this transformation using two custom CUDA kernels. The first kernel computes the column-wise sum of the weight matrix with fully coalesced memory accesses. The second kernel performs the dot product between the input vector and the reduced weight vector, while simultaneously applying division and scaling. These operations are fused into a single kernel to avoid intermediate writes to global memory.

To further improve performance, the dot product kernel leverages vectorized memory access using \texttt{float4} loads, maximizing memory bandwidth utilization. A shared-memory tree reduction is used within each thread block to accumulate partial sums efficiently, reducing synchronization overhead and avoiding expensive global atomics.

This case demonstrates how \ours systematically combines algorithmic simplification with kernel fusion and low-level CUDA optimizations to collapse a multi-operator computation graph into a small number of highly efficient kernels.


\begin{figure}[htbp]
  \centering
  \includegraphics[width=0.8\linewidth]{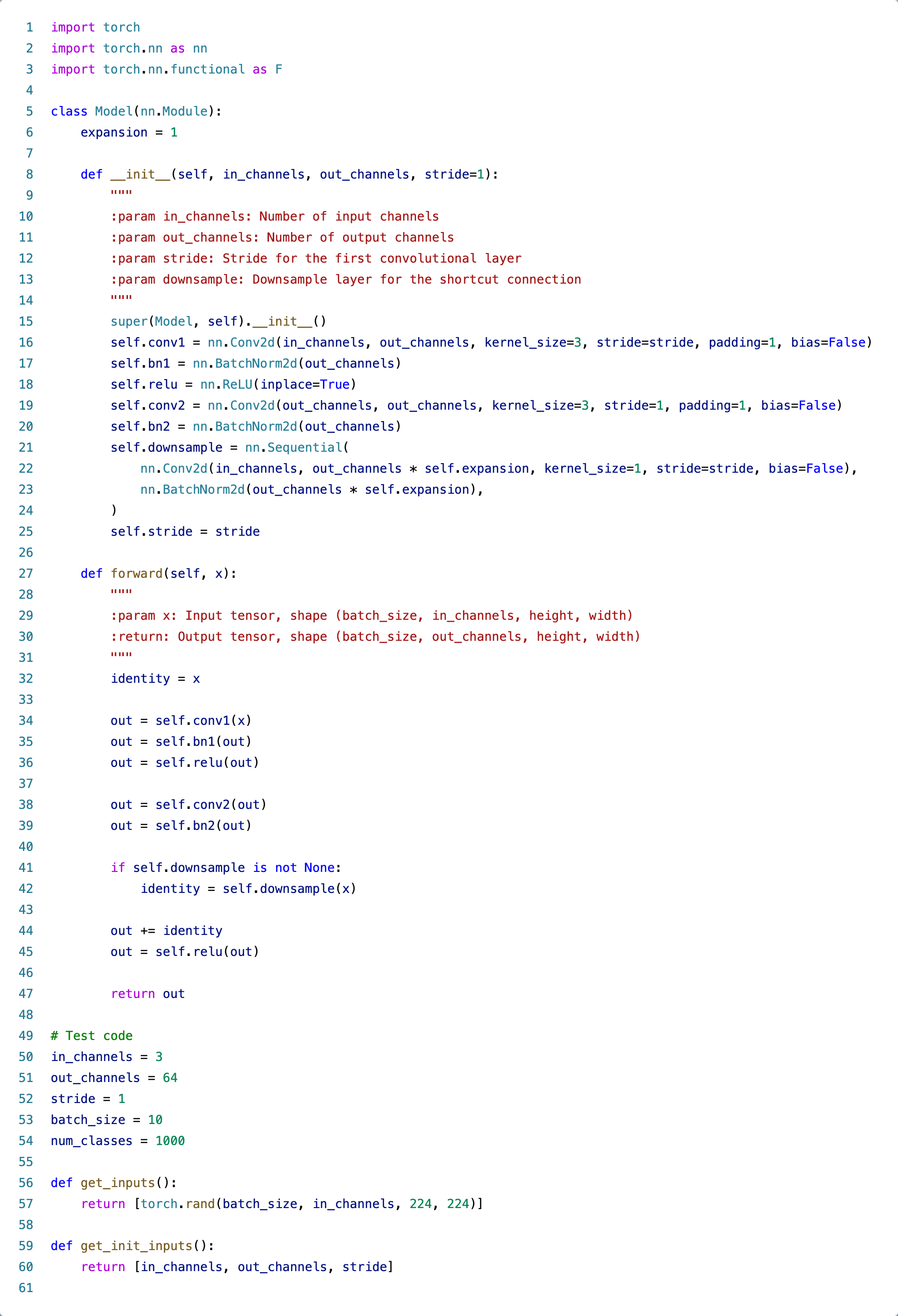}
  \caption{Reference operator for Resnet BasicBlock (Case~\ref{case:3}).}
  \label{fig:level3_1_248_p1}
\end{figure}

\begin{figure}[htbp]
  \centering
  \includegraphics[width=0.8\linewidth]{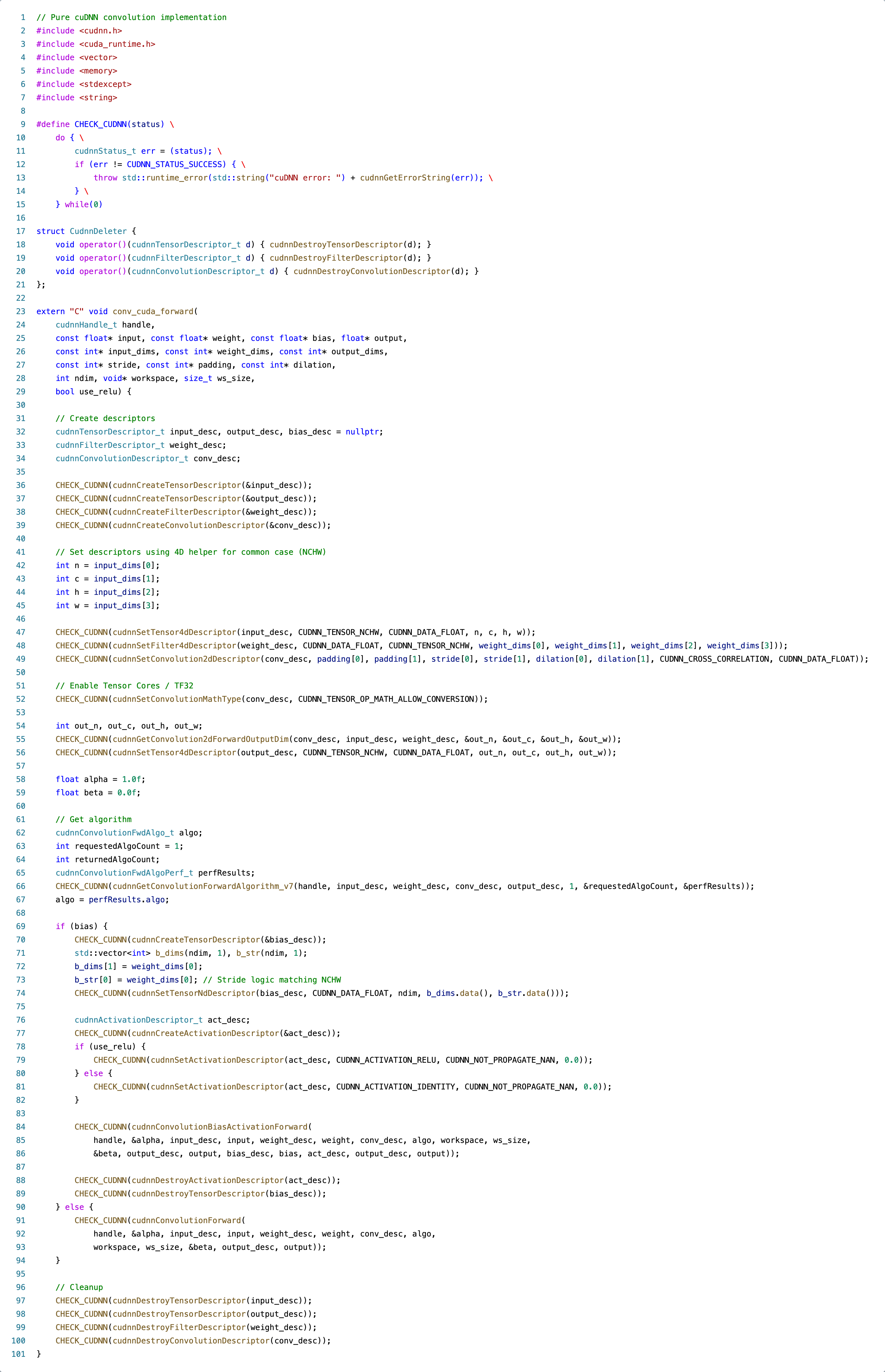}
  \caption{\texttt{cuDNN} convolution implementation, part~1 (Case~\ref{case:3}).}
  \label{fig:level3_1_248_p2}
\end{figure}

\begin{figure}[htbp]
  \centering
  \includegraphics[width=0.8\linewidth]{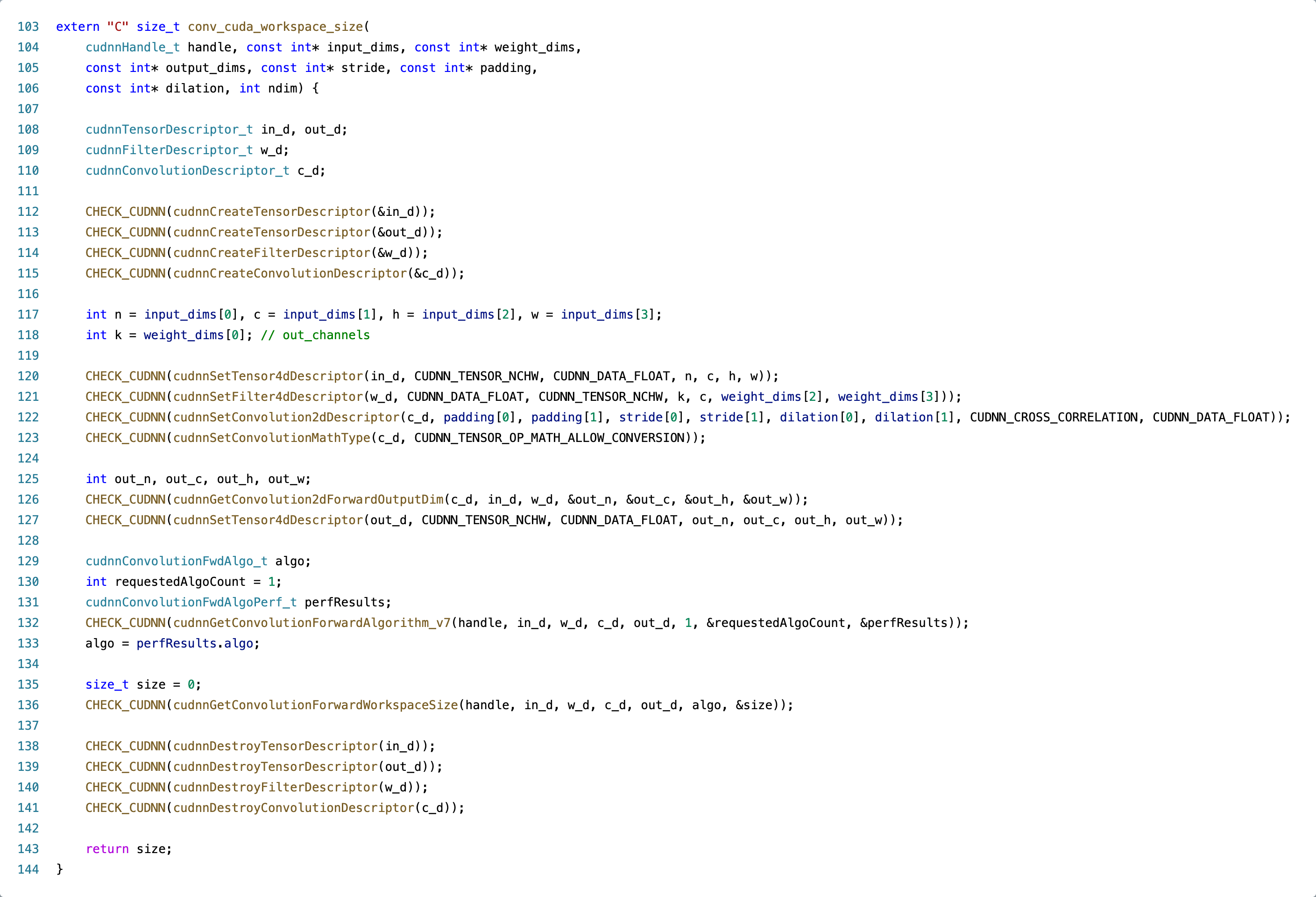}
  \caption{\texttt{cuDNN} convolution implementation, part~2 (Case~\ref{case:3}).}
  \label{fig:level3_1_248_p3}
\end{figure}


\begin{figure}[htbp]
  \centering
  \includegraphics[width=0.8\linewidth]{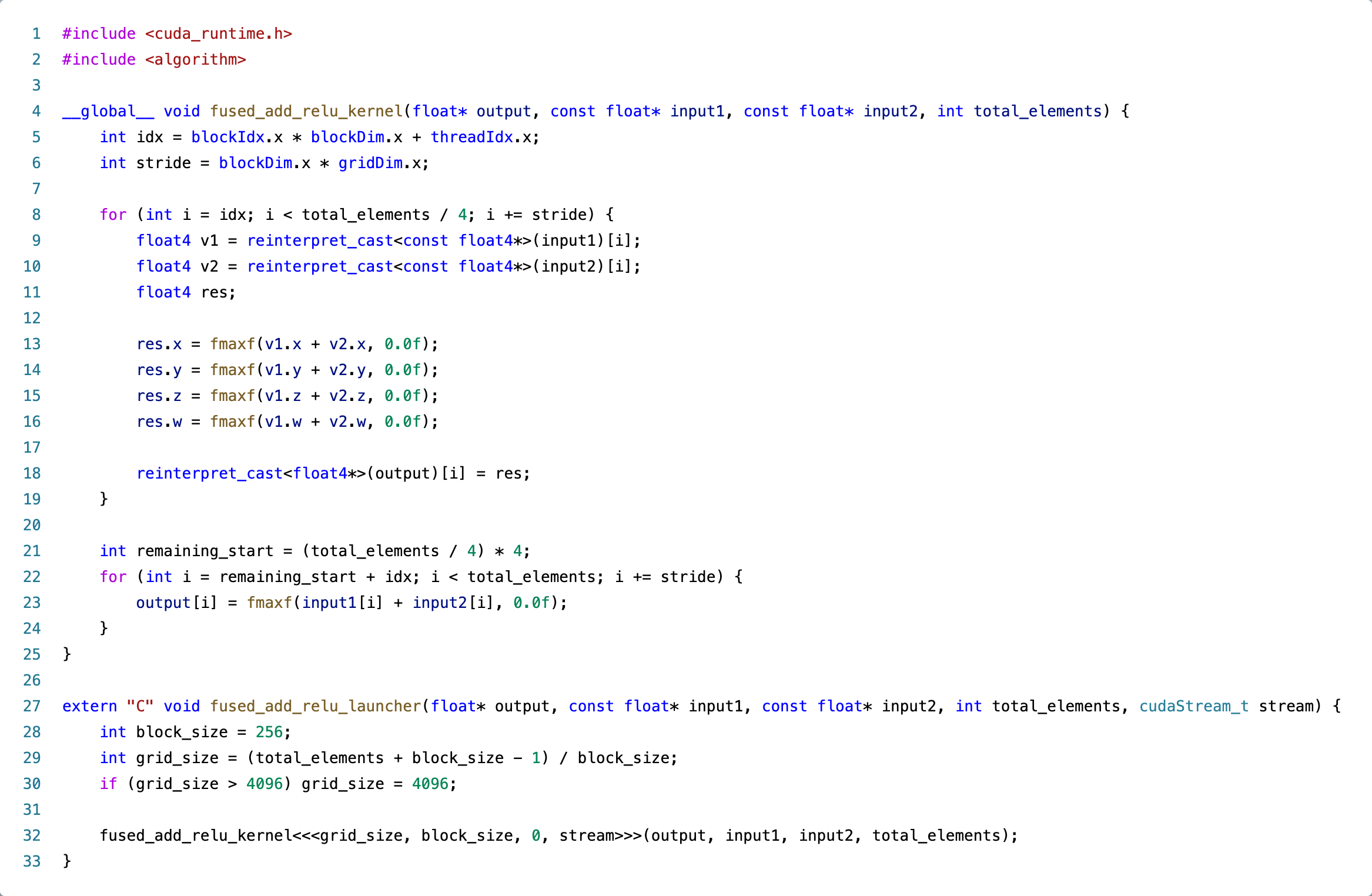}
  \caption{Fused \texttt{add-relu} kernel implementation (Case~\ref{case:3}).}
  \label{fig:level3_1_248_p5}
\end{figure}


\begin{figure}[htbp]
  \centering
  \includegraphics[width=0.6\linewidth]{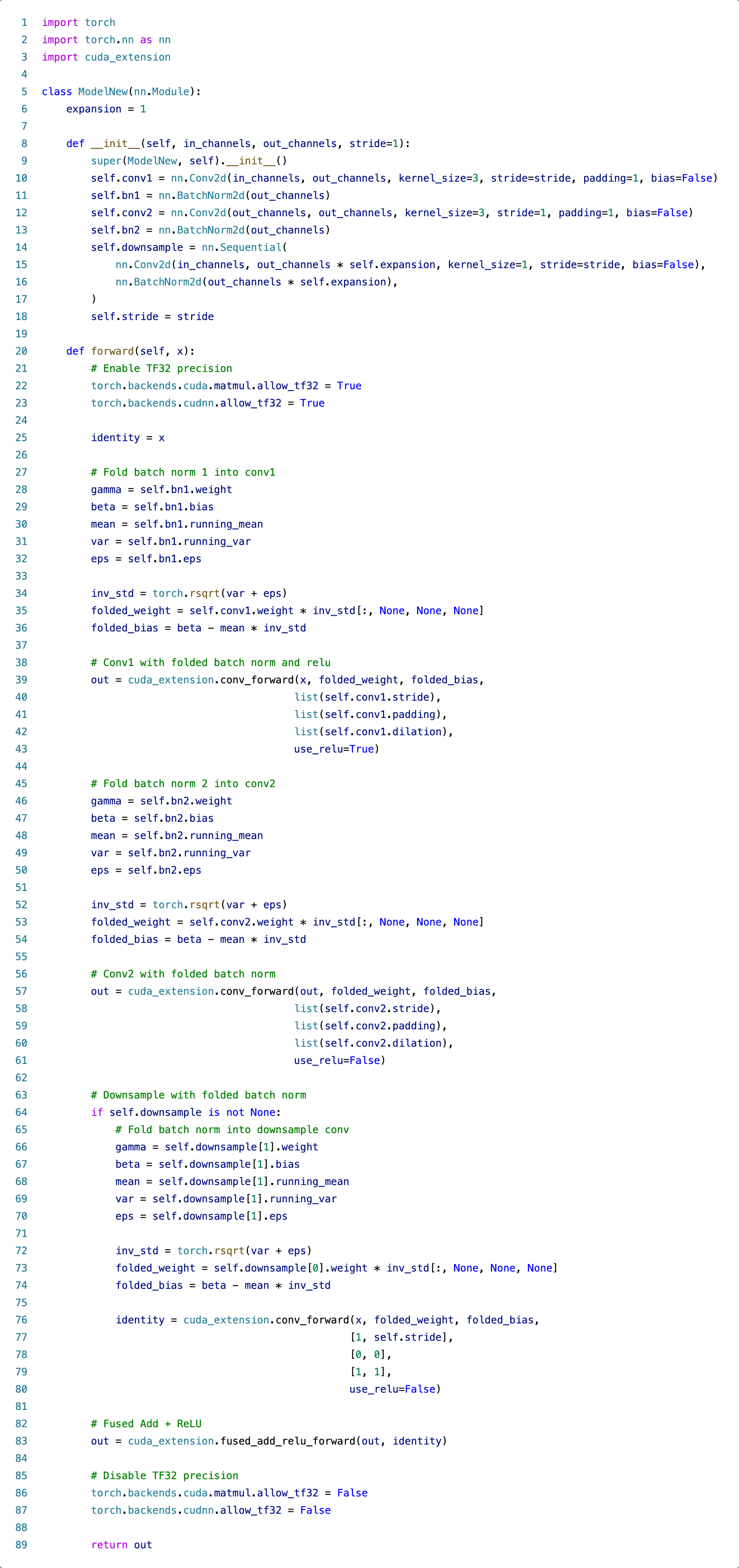}
  \caption{Custom operator for Resnet BasicBlock (Case~\ref{case:3}).}
  \label{fig:level3_1_248_p7}
\end{figure}

\subsection{Example of Level~3: ResNet BasicBlock}
\label{case:3}

The third case studies a ResNet BasicBlock, representative of realistic deep learning workloads composed of convolutions, normalization, nonlinearities, and residual connections. The reference implementation follows the standard PyTorch execution path, resulting in multiple kernel launches for convolution, batch normalization, bias addition, activation, and residual addition. Figures~\cref{fig:level3_1_248_p1,fig:level3_1_248_p2,fig:level3_1_248_p3,fig:level3_1_248_p5,fig:level3_1_248_p7} present the reference ResNet BasicBlock, the optimized cuDNN-based convolution, the fused add--ReLU kernel, and the resulting custom operator. The Python/C++ binding is omitted for brevity. This custom operator achieves 3.59$\times$ speed-up versus Torch Compile.

\ours applies several complementary techniques. First, it manually folds the BatchNorm parameters into the preceding convolution weights and bias in the Python model, eliminating the BatchNorm operator entirely at inference time. This reduces both kernel launches and memory accesses while preserving numerical equivalence.

Second, by using \texttt{cudnnConvolutionBiasActivationForward}, \ours modifies the custom cuDNN convolution wrapper to enable convolution, bias addition, and ReLU activation to be executed within a single cuDNN kernel. This further reduces kernel launch overhead and improves data locality. We explicitly enable TF32 computation for cuDNN and matrix multiplication operations, allowing the implementation to leverage Tensor Cores on Hopper GPUs.

\ours also explores switching the data layout from NCHW to NHWC to better align with Tensor Core requirements. However, the required layout conversions introduced significant overhead, offsetting the potential gains. As a result, the final implementation retains the NCHW layout while using NCHW-compatible cuDNN APIs.

Finally, \ours fuses the residual addition and the final ReLU activation into a single custom CUDA kernel, replacing two separate element-wise operators. This kernel computes the element-wise sum of the main and residual branches and immediately applies the ReLU function.

Together, these optimizations reduce the number of kernel launches, improve arithmetic intensity, and better exploit hardware acceleration features. This case highlights \ours’s ability to integrate high-level graph transformations with library-level fusion and custom kernel design in complex, real-world neural network blocks.

\section{Limitations}
Our study has two main limitations. First, we do not compare \ours against more sophisticated compiler frameworks such as TVM. While these systems can potentially provide stronger baselines, they are difficult to integrate into a large-scale RL training loop with thousands of rollouts due to their substantial tuning overhead and complex deployment requirements; we therefore focus on \texttt{torch.compile} as a widely adopted, training-friendly baseline. Second, our training pipeline relies on a large GPU pool with process-level isolation, which incurs considerable computational and engineering cost. This reliance on massive GPU resources may limit accessibility for broader research community, and we leave exploring more resource-efficient training strategies as important directions for future work.

\end{document}